\documentclass[ijoc,nonblindrev]{article} 

\usepackage{natbib}
 \bibpunct[, ]{(}{)}{,}{a}{}{,}%
 \def\bibfont{\small}%
\usepackage{subfig}
\usepackage{blindtext}
\usepackage{amsmath}
\usepackage{graphicx}

\begin{document}




\title{Iterative Rule Extension for Logic Analysis of Data: an MILP-based heuristic to derive interpretable binary classification from large datasets}

\author{Marleen Balvert\\Department of Econometrics \& Operations Research, Tilburg University,\\ Tilburg, The Netherlands\\ m.balvert@tilburguniversity.edu}
\date{}

\maketitle

\begin{abstract}
Data-driven decision making is rapidly gaining popularity, fuelled by the ever-increasing amounts of available data and encouraged by the development of models that can identify beyond linear input-output relationships. Simultaneously the need for interpretable prediction- and classification methods is increasing, as this improves both our trust in these models and the amount of information we can abstract from data. An important aspect of this interpretability is to obtain insight in the sensitivity-specificity trade-off constituted by multiple plausible input-output relationships. These are often shown in a receiver operating characteristic (ROC) curve. These developments combined lead to the need for a method that can abstract complex yet interpretable input-output relationships from large data, i.e. data containing large numbers of samples and sample features. Boolean phrases in disjunctive normal form (DNF) are highly suitable for explaining nonlinear input-output relationships in a comprehensible way. Mixed integer linear programming (MILP) can be used to abstract these Boolean phrases from binary data, though its computational complexity prohibits the analysis of large datasets. This work presents IRELAND, an algorithm that allows for abstracting Boolean phrases in DNF from data with up to 10,000 samples and sample characteristics. The results show that for large datasets IRELAND outperforms the current state-of-the-art and can find solutions for datasets where current models run out of memory or need excessive runtimes. Additionally, by construction IRELAND allows for an efficient computation of the sensitivity-specificity trade-off curve, allowing for further understanding of the underlying input-output relationship. 
\end{abstract}%




%


\section{Introduction}
Over the past decade the field of machine learning and artificial intelligence (AI) has seen major developments alongside a tremendous increase in popularity among academics, industry and the general public. Supervised machine learning models, which are among the most frequently used approaches in AI, aim to learn the relationship between input features and an output feature or class label. Examples are training a model to ``read'' handwritten digits from images, recommending a new video for streaming service customers based on their viewing history, or recommending a treatment for cancer patients based on their tumor's genetic characteristics. Over the past decade the development of supervised learning models was mainly focused on improving prediction or classification accuracy. Many of the developed methods are, to various degrees, black box methods: while they yield high prediction accuracy, the input-output relationship that the models identify and base their predictions on is difficult to comprehend or even invisible to humans. 

The interpretability of machine learning methods is essential for their acceptance for several reasons \citep{doshi2017towards,molnar2020interpretable}. First, when decisions are made that impact people's lives, users need to understand why a model makes certain predictions in order to trust them. This particularly holds in the case of medical applications. Second, for several applications the relationship between input data and predictions is more important than the predictions themselves. For example, when developing medication one needs to understand the biological processes that cause a disease and should be targeted by the drug. Analyzing bioinformatics data with machine learning models that do not only provide predictions of drug response but also give insight in the underlying input data-prediction relationship can play an important role. Third, the General Data Protection Regulation of the EU requires that a data subject has the right to explanation when decisions affecting them are made using automated models \citep{eu-gdpr}. These motivations have lead to an increased interest in developing interpretable machine learning models \citep[and references therein]{molnar2020interpretable}.

In the case of predicting a binary class from binary input data, the focus of this work, Boolean phrases are very well suited for prediction while providing an interpretable and comprehensible input-output relationship \citep{lakkaraju2016interpretable}. This work focuses on identifying a Boolean phrase in disjunctive normal form (DNF), which is an OR combination of AND clauses. For example, the following is a Boolean statement in DNF: ``if $(X_{n,5}=1$ AND $X_{n,12}=1)$ OR $(X_{n,2}=1$ AND $X_{n,3}=1$ AND $X_{n,25}=1)$ OR $(X_{n,22}=1)$, then sample $n$ is predicted to be in class 1, else it is predicted to be in class 0'', where $X\in \{0,1\}^{N\times J}$ denotes the input matrix for a dataset with $N$ samples and $J$ features. This data format is motivated by applications in medical genetics, where combinations of genetic variants lead to disease or drug resistance. Individuals either do or do not have the considered genetic characteristics, represented in the matrix $X$, and they do or do not have a certain personal trait, represented by the binary class. Note that categorical and continuous input data can be transformed into binary data \citep{boros1997logical}.

Identifying Boolean phrases in DNF for classification from binary data has been an active research topic in learning theory, especially since \cite{valiant1984theory} posed the question whether DNF rules were efficiently learnable from data. The work in this field has focused on developing solution algorithms and providing the corresponding complexity bounds for the noiseless setting \citep{bshouty1996subexponential,tarui1999learning,klivans2004learning}. No efficient algorithm was found, and recently \cite{daniely2016complexity} showed that learning DNF rules from data is hard.


Integer programming has been shown to be a suitable method for identifying Boolean phrases in DNF from binary data \citep{hauser2010disjunctions,hammer2006logical,chang2012integer,malioutov2013exact,wang2015learning,knijnenburg2016logic,dash2018boolean}. The approach was previously termed the Logical Analysis of Data. While existing approaches work well for datasets of limited size, novel solution algorithms to solve large instances are needed: with the current rapid increase in data collection efforts and skills, datasets containing millions of features for thousands of samples are now available for analysis. The number of binary variables included in the integer program strongly increases with the number of samples, the number of features per sample and the number of AND clauses included. As a result, the integer program cannot be applied to these large datasets. \cite{dash2018boolean} have taken the first step in overcoming this issue: they  developed a column generation approach where in each iteration a new AND clause is generated, forming a new column in the overall problem. In order to do so, while others minimize the classification error, \cite{dash2018boolean} need to minimize the Hamming loss defined as the number of false negatives plus the number of AND clauses satisfied by each of the controls, summed over the controls. While this resolves the issue of the increase in the number of binary variables with the number of AND clauses, the effect of the number of samples and features on the complexity partially remains as the sub problem is large for a large number of samples and features.

This work presents a solution algorithm that allows for solving the mixed integer linear program (MILP) for datasets with a large number of samples and features. The algorithm is termed IRELAND, Iterative Rule Extension for Logical Analysis of Data, and breaks up the MILP into smaller problems. Similar to \cite{malioutov2013exact} and \cite{dash2018boolean}, the algorithm uses a sub problem to generate a set of promising AND clauses. From this set the master problem selects those AND clauses that, when combined through OR statements, yield the best accuracy. Each sub problem considers only a subset of the samples, containing all controls and only those cases that have not been classified as cases in the previous solution. As such the sub problem focuses on adding an AND clause that, when added to the Boolean phrase of the previous solution, increases the number of true positives while limiting the increase in the number of false positives.

Besides achieving maximum accuracy, users of classification models are often interested in the trade-off between sensitivity and specificity. 
By construction IRELAND allows for easy computation of the sensitivity-specificity trade-off curve. When directly optimizing the original MILP one can only obtain information on this trade-off by solving an MILP where the objective function is to maximize sensitivity while placing a constraint on specificity or vice versa. Varying the lower bound on specificity (or sensitivity) provides the trade-off curve between sensitivity and specificity. This means that a computationally heavy MILP needs to be solved several times. IRELAND on the other hand naturally accommodates the analysis of the sensitivity-specificity trade-off, as it generates a large pool of promising AND clauses. The master problem, which is now maximizing sensitivity while constraining the specificity, can then be solved several times for various lower bounds on specificity, selecting combinations of AND clauses that provide different trade-offs between sensitivity and specificity.

This paper contains four contributions. First, several formulations of the MILP are compared based on runtime and objective value for small datasets. Second, an algorithm is introduced, called IRELAND, that allows for solving problems for the Logical Analysis of Data for datasets with more than 1,000 samples and features. Third, rules of thumb are provided for which datasets IRELAND gives the best performance, and for which datasets the original MILP  or the model proposed by \cite{dash2018boolean}  perform best. Fourth, IRELAND enables the efficient construction of the sensitivity-specificity trade-off curve, a useful feature in many real-world applications. All code and datasets will be made publically available upon acceptance of the manuscript.

\section{Methods}\label{sec:Methods}
The MILP that abstracts Boolean phrases from data can be formulated in several ways. In Section \ref{sec:meth:MILPformulation} the formulations are provided and compared. As all MILP formulations are limited in the size of the data that they can process, Section \ref{sec:meth:IRELANDalgorithm} presents the proposed algorithm IRELAND. An extension to generating the sensitivity-specificity trade-off curve is presented in Section \ref{sec:meth:Pareto}. Section \ref{sec:meth:Datasets} explains how test data was generated.

\subsection{Mixed integer linear programming formulation}\label{sec:meth:MILPformulation}
Let $\mathcal{N} = \{1,...,N\}$ denote the set of all samples and $\mathcal{J}=\{1,...,J\}$ the set of features. Let $X \in \{0,1\}^{N\times J}$ denote the feature matrix where $X_{nj} = 1$ if sample $n$ has characteristic $j$ and 0 otherwise. Let $y \in \{0,1\}^N$ denote the class vector, where $y_n=1$ if sample $n$ is a case and $y_n=0$ if sample $n$ is a control. Based on $X$ and $y$, the model will identify Boolean phrases in DNF that predict a sample's class from input information $X_{n\cdot}$. The variables $\hat{y}\in\{0,1\}^N$ represent the predicted class for each sample.


The MILP aims to find a Boolean phrase in DNF that yields the best balanced prediction accuracy:
\begin{align*}
    \min_{\hat{y}} \sum_{n \in \mathcal{N}_0} w_n\hat{y}_n - \sum_{n \in \mathcal{N}_1} w_n(1 - \hat{y}_n),
\end{align*}
where $\mathcal{N}_0$ denotes the set of controls and $\mathcal{N}_1$ the set of cases. Weights $w_n$ account for an inbalance between the number of cases and controls in the dataset, and are defined as:
\begin{equation*}
    w_n = \begin{cases}
         \frac{|\mathcal{N}_1|}{N} & \text{if }n \in \mathcal{N}_0 \\
         \frac{|\mathcal{N}_0|}{N} & \text{if }n \in \mathcal{N}_1
    \end{cases}
\end{equation*}
Alternatively, one could minimize the Hamming loss \citep{lakkaraju2016interpretable,dash2018boolean}, which is defined as the number of incorrectly classified cases plus the number of AND clauses that each control satisfies. Let $t_{nk} \in \{0,1\}$ for $n \in \mathcal{N}, k \in \mathcal{K}$ be an auxiliary variable that denotes whether sample $n$ satisfies AND clause $k$, that is:
\begin{align*}
    t_{nk} = 
    \begin{cases}
        1 & \text{if } X_{nj} = 1 \ \forall \ j \text{ that is in AND clause }k \\
        0 & \text{otherwise.} \\
    \end{cases}
\end{align*}
The Hamming loss can then be computed as:
\begin{align*}
    \sum_{n \in \mathcal{N}_0} \sum_{k \in \mathcal{K}} w_n t_{nk} + \sum_{n \in \mathcal{N}_1} w_n(1 - y_n),
\end{align*}
where $\mathcal{K}$ denotes the set of AND clauses.

The AND clauses and OR combinations of AND clauses are modeled by separate constraints. OR rules can be represented by two different sets of constraints. The following set of constraints ensures that, for given $n$, $\hat{y}_n=1$ if and only if $\exists k: t_{nk}=1$ \citep{knijnenburg2016logic}:
\begin{subequations}
    \begin{align}
        \hat{y}_n - \sum_{k \in K} t_{nk} & \leq 0 & \forall n \in \mathcal{N}_1 \quad  \label{eq:OR1a} \\
        K\hat{y}_n - \sum_{k \in K} t_{nk} & \geq 0 & \forall n \in \mathcal{N}_0 \quad  \label{eq:OR1b} \\
        \hat{y}_n \in \{0,1\} \quad & & \forall n \in \mathcal{N}. \nonumber
    \end{align}
    \label{eq:OR1}
\end{subequations}
Together with an objective function that minimizes (maximizes) $\hat{y}_n$ for $n \in \mathcal{N}_0$ ($n \in \mathcal{N}_1$) these constraints yield the correct values for $\hat{y}_n$. Constraints \eqref{eq:OR1} are equivalent to:
\begin{subequations}
    \begin{align}
        \hat{y}_n - \sum_{k \in \mathcal{K}} t_{nk} &\leq 0 & \forall n \in \mathcal{N}_1  \label{eq:OR2a} \\
        - \hat{y}_n + t_{nk} & \leq 0 & \forall k \in \mathcal{K}, \quad \forall n \in \mathcal{N}_0  \label{eq:OR2b} \\
        \hat{y}_n \in \{0,1\} & & \forall n \in \mathcal{N}. \nonumber
    \end{align}
    \label{eq:OR2}
\end{subequations}
Although the feasible regions described by constraint sets \eqref{eq:OR1} and \eqref{eq:OR2} are identical, their relaxations, where the integrality constraint on $\hat{y}_n$ is replaced by $\hat{y}_n \in [0,1]$, are not, see Appendix A. For the relaxation the feasible region defined by constraints \eqref{eq:OR2} is a subset of the feasible region described by constraints \eqref{eq:OR1} (Appendix A).

AND clauses can be represented in two different ways as well. Let $s_{kj} \in \{0,1\}$ for $k \in \mathcal{K}, j \in \mathcal{J}$ denote the vector of decision variables indicating whether feature $j$ is included in the $k^\text{th}$ AND clause. The following set of constraints enforce $t_{nk}=0$ if and only if there exists no $j$ such that $s_{kj}=1$ and $X_{nj}=0$ \citep{knijnenburg2016logic}:
\begin{subequations}
    \begin{align}
    J \cdot t_{nk} + \sum_{j \in \mathcal{J}} (1-X_{nj})s_{kj} & \leq J & \forall k \in \mathcal{K}, \ \forall n \in \mathcal{N}_1  \label{eq:AND1a} \\
    t_{nk} + \sum_{j \in \mathcal{J}} (1-X_{nj})s_{kj} & \geq 1 & \forall k \in \mathcal{K}, \ \forall n \in \mathcal{N}_0  \label{eq:AND1b} \\
    s_{jk}, t_{nk} \in \{0,1\} & & \forall j \in \mathcal{J}, \ \forall k \in \mathcal{K}, \ \forall n \in \mathcal{N}. \nonumber
    \end{align}
    \label{eq:AND1}
\end{subequations}
Note that these constraints hold because $t_{nk}$ is minimized for $n \in \mathcal{N}_0$ and maximized for $n \in \mathcal{N}_1$. The following alternative set of equations yields the same feasible region:
\begin{subequations}
    \begin{align}
    t_{nk} + (1-X_{nj})s_{kj} &  \leq 1 & \forall j \in \mathcal{J}, \ \forall k \in \mathcal{K}, \ \forall n \in \mathcal{N}_1 \label{eq:AND2a} \\
    t_{nk} + \sum_{j \in \mathcal{J}} (1-X_{nj})s_{kj} & \geq 1 & \forall k \in \mathcal{K}, \ \forall n \in \mathcal{N}_0. \label{eq:AND2b} \\
    s_{jk}, t_{nk} \in \{0,1\} & & \forall j \in \mathcal{J}, \ \forall k \in \mathcal{K}, \ \forall n \in \mathcal{N}. \nonumber
    \end{align}
    \label{eq:AND2}
\end{subequations}
\noindent For $t_{nk}, s_{kj} \in \{0,1\}$ constraints \eqref{eq:AND1} are equivalent to \eqref{eq:AND2}. However, when the integrality constraints are relaxed to $t_{nk}$, $s_{jk} \in [0,1]$, the polyhedron defined by \eqref{eq:AND2} is a subset of the polyhedron defined by equations \eqref{eq:AND1}, see Appendix A.

Eight different MILPs can be formulated to abstract Boolean phrases in DNF from data by combining one of the objective functions with one of the formulations for the OR rules and one of the formulations for AND clauses as defined above. Note that when the objective is to minimize the Hamming loss, the variables $\hat{y}_n$ for $n \in \mathcal{N}_0$ are redundant, as are constraints \eqref{eq:OR1b} and \eqref{eq:OR2b}. Since constraints \eqref{eq:OR1a} and \eqref{eq:OR2a} are identical, only six formulations remain, see Table \ref{tab:overviewModels} for an overview. The models in this manuscript use an additional constraint that bounds the number of features in an AND clause by $M$:
\begin{align}
    \sum_{j \in \mathcal{J}} s_{kj} \leq M & \qquad \forall k \in \mathcal{K}. \label{eq:bigM}
\end{align}

For each formulation some of the binary variables can be relaxed without altering the optimal solution. As an example, consider the MILP that minimizes classification accuracy such that \eqref{eq:OR1}, \eqref{eq:AND2} and \eqref{eq:bigM} are satisfied:
\begin{align*}
    \max \quad & \sum_{n \in \mathcal{N}_1} w_n \hat{y}_n + \sum_{n \in \mathcal{N}_0} w_n (1-\hat{y}_n) \\
    \text{s.t.} \quad & t_{nk} + (1-X_{nj})s_{kj} \leq 1 & \forall j \in \mathcal{J}, \ \forall k \in \mathcal{K}, \ \forall n \in \mathcal{N}_1 \\
    & t_{nk} + \sum_{j \in \mathcal{J}} (1-X_{nj})s_{kj} \geq 1 & \forall k \in \mathcal{K}, \ \forall n \in \mathcal{N}_0 \\
    & \hat{y}_n - \sum_{k \in K} t_{nk} \leq 0 & \forall n \in \mathcal{N}_1   \\
    & K\hat{y}_n - \sum_{k \in K} t_{nk} \geq 0 & \forall n \in \mathcal{N}_0  \\
    & \sum_{j \in \mathcal{J}} s_{kj} \leq M & \forall k \in \mathcal{K} \\
    & \hat{y}_n, \ s_{jk}, \ t_{nk} \in \{0,1\} \quad & \forall n \in \mathcal{N}.
\end{align*}
First note that the lower bound on $t_{nk}$ determined by constraint \eqref{eq:AND2b}, i.e. $t_{nk} \geq 1-\sum_{j \in \mathcal{J}} (1-X_{nj})s_{kj}$, is always integer. Since $t_{nk}$ is minimized for $n\in \mathcal{N}_0$, it will become integer, hence the integrality constraint on $t_{nk}, n \in \mathcal{N}_0$, can be relaxed. 
Following a similar reasoning, one can relax the integrality constraint on $\hat{y}_n$ for $n \in \mathcal{N}_1$. For $t_{nk}$, $n \in \mathcal{N}_1$, integrality of the optimal solution cannot be guaranteed when relaxing the integrality constraint on these variables. For example, suppose $s_{k1}=s_{k2}=1$ for some $k$, then for a given $n \in \mathcal{N}_1$, $t_{nk}=0.5$ is feasible and allows $\hat{y}_n$ to be equal to 1. 

Each of the six models has a different subset of variables that can be relaxed from binary to the interval $[0,1]$. An overview of the models considered in this work, their constraints and objective, the number of constraints and the number of binary and continuous variables is given in Table \ref{tab:overviewModels}. Note that $(BP4)$ can be simplified to reduce the total number of variables and constraints by combining constraints \eqref{eq:AND2a} and \eqref{eq:OR2b} into one:
\begin{align}
    \hat{y}_n \geq 1 - \sum_{j \in \mathcal{J}} (1-X_{nj})s_{kj} \quad \forall k \in \mathcal{K}, \ \forall n \in \mathcal{N}_0. \label{eq:ANDORcombo}
\end{align}
This makes the variables $t_{nk}$ for $n \in \mathcal{N}_0$ obsolete.

\begin{table}
    \centering
    \caption{Six MILP formulations that abstract Boolean phrases in DNF from binary data. $N_0$ denotes the number of controls, $N_1$ the number of cases, and $N=N_0+N_1$. $K$ denotes the number of AND clauses included in the model, and $J$ the number of features.}\label{tab:overviewModels}
    \begin{tabular}{llccccc}
        \hline
        Model & Objective & \multicolumn{2}{c}{Constraints} &  Number of constraints & \multicolumn{2}{c}{Number of Variables} \\
        & & OR & AND & & Continuous & Binary \\
        \hline
         $(BP1)$ & Accuracy & \eqref{eq:OR1} & \eqref{eq:AND1} & $K+NK+N$ & $N_0K+N_1$ & $JK+N_1K+N_0$ \\
         $(BP2)$ & Accuracy & \eqref{eq:OR1} & \eqref{eq:AND2} & $K+N_0K+N_1JK+N$ & $NK+N_1$ & $JK+N_0$ \\
         $(BP3)$ & Accuracy & \eqref{eq:OR2} & \eqref{eq:AND1} & $K+NK+N_0K+N_1$ & $NK+N$ & $JK$ \\
         $(BP4)$ & Accuracy & \eqref{eq:OR2} & \eqref{eq:AND2} & $K+N_0K+N_1+N_1JK$ & $N_1K+N$ & $JK$ \\
         $(BP5)$ & Hamming loss & \eqref{eq:OR1a} & \eqref{eq:AND1} & $K+NK+N_1$ & $N_0K+N_1$ & $JK+N_1K$ \\
         $(BP6)$ & Hamming loss & \eqref{eq:OR1a} & \eqref{eq:AND2} & $K+N_0K+N_1JK+N_1$ &  $NK+N_1$ & $JK$ \\
         \hline
    \end{tabular}
\end{table}

\subsection{IRELAND: a solution algorithm}\label{sec:meth:IRELANDalgorithm}

The complexity of the MILP is due to its large number of binary variables (Table \ref{tab:overviewModels}). As can be seen from Appendix B, an increase in the number of samples $N$, the number of features $J$ and the number of included AND clauses $K$ all lead to an increase in solution time. To overcome the computational burden arising from large data, this work presents the solution algorithm IRELAND: Iterative Rule Extension for Logical ANalysis of Data. The idea behind IRELAND is to break up the problem into sub problems that contain only a subset of the variables, mostly limiting $N$ and $K$. The sub problems together generate a large pool of AND clauses with various levels of sensitivity and specificity, in preparation for generating the trade-off curve. 

\begin{figure}
	\centering
	\includegraphics[width=\textwidth]{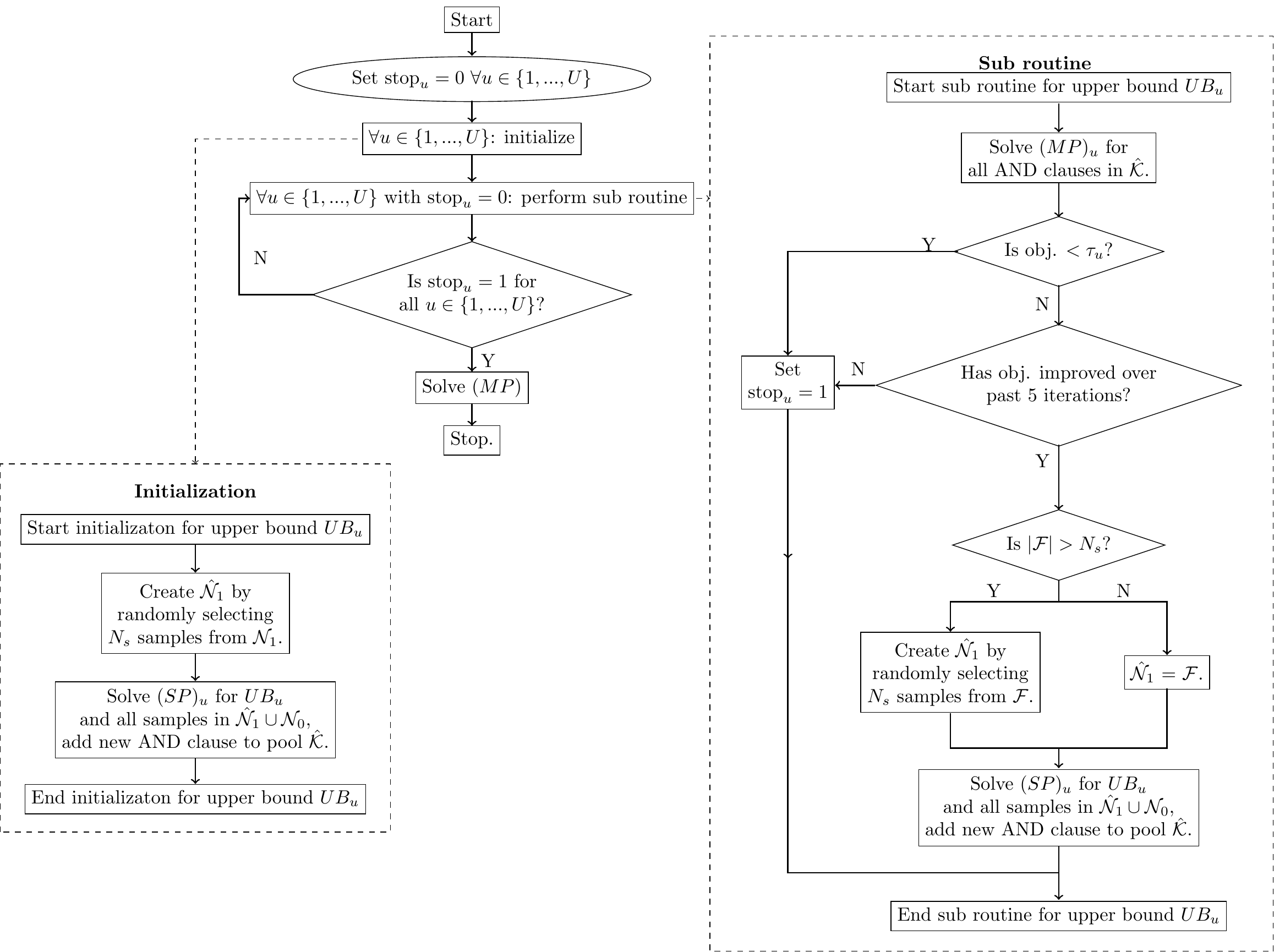}
	\caption{Flow chart of IRELAND. $\{UB_1,...,UB_U\}$ is an a priori chosen set of upper bounds on the number of false positives. $\mathcal{N}_0$ and $\mathcal{N}_1$ are the sets of controls and cases, respectively, $\mathcal{F}$ is the set of false negatives and $N_s$ is an a priori chosen size for the subset $\hat{\mathcal{N}}$. $\hat{\mathcal{K}}$ denotes the pool of AND clauses that is generated iteratively. $\{\tau_1,...,\tau_U\}$ are objective function values for $(MP)_1$,...,$(MP)_U$ that are sufficient for the algorithm to stop.}
	\label{fig:IRELAND}
\end{figure}

The algorithm is summarized in Figure \ref{fig:IRELAND}. IRELAND consists of two components: the initialization where an initial pool of AND clauses is generated (left part of Figure \ref{fig:IRELAND}), and the sub routine that iteratively generates AND clauses (right part of Figure \ref{fig:IRELAND}). IRELAND uses three MILPs, namely a sub problem for the initialization and sub routine, a master problem for the sub routine and an overall master problem. Each of the MILPs uses constraints \eqref{eq:OR1} and \eqref{eq:AND1}, see Section \ref{sec:res:MILP} for a motivation of this choice. Details of the initialization, the sub routine and the three MILPs are given below.

\emph{The sub problem} Both the initialization and the sub routine make use of the sub problem. Every time the sub problem is solved an AND clause is generated and added to the pool $\mathcal{\hat{K}}$. The sub problem generates a single AND clause by maximizing the number of true positives, while restricting the number of false positives to be at most $UB_u$, $u \in \{1,...,U\}$:
\begin{align}
	(SP)_u \qquad \max_{s,\hat{y}} \qquad & \sum_{n \in \hat{\mathcal{N}}_1} \hat{y}_n \nonumber \\
	\text{s.t.} \qquad & \sum_{n \in \mathcal{N}_0} \hat{y}_n \leq UB_u & 
	\nonumber \\
	& J \cdot \hat{y}_{n} + \sum_{j \in \mathcal{J}} (1-X_{nj})s_{j} \leq J & \forall n \in \hat{\mathcal{N}}_1 \nonumber \\
	& \hat{y}_{n} + \sum_{j \in \mathcal{J}} (1-X_{nj})s_{j} \geq 1 & \forall n \in \mathcal{N}_0  \nonumber \\
	& \sum_{j \in \mathcal{J}} s'_j \cdot s_j + \sum_{j \in \mathcal{J}} (1-s'_j)(1-s_j) \leq J-1 & \forall s' \in \mathcal{\hat{K}} \label{eq:SP:nodoubleclauses}\\
	& \sum_{j \in \mathcal{J}} s_j \leq M \nonumber \\ 
	& s_{j}, \hat{y}_{n} \in \{0,1\},\nonumber 
\end{align}
where $\hat{y}$, $s$ and $\mathcal{\hat{K}}$ are as before. Constraint \eqref{eq:SP:nodoubleclauses} ensures that the newly generated AND clause is different from all AND clauses that are in $\mathcal{\hat{K}}$ already, where $s'$ is a parameter representing the AND clauses in $\mathcal{\hat{K}}$. Note that $(SP)_u$ is solved for all samples in $\mathcal{N}_0 \cup \hat{\mathcal{N}}_1$, where $\hat{\mathcal{N}}_1 \subset \mathcal{N}_1$.

\emph{The initialization} In the initialization phase $(SP)_u$ is solved for all upper bounds in the predefined set $\{UB_1,...,UB_U\}$. Even though the sub problem $(SP)_u$ only solves for a single AND clause, it still takes a large amount of time when $N$ is large. Therefore, a random subset of $\hat{\mathcal{N}}_1 \subset \mathcal{N}_1$ of size $N_s$ is selected. Each upper bound $UB_u$ contributes one AND clause to the initial pool $\hat{\mathcal{K}}$. 

\emph{The sub routine master problem} In every call of the sub routine, a slight modification of the master problem is solved. For a given upper bound $UB_u$, the sub routine master problem chooses those AND clauses from the pool $\hat{\mathcal{K}}$ that maximize the number of true positives while limiting the number of false positives to be at most $UB_u$:
\begin{align}
	(MP)_u \qquad \max_{q,\hat{y}} \qquad & \sum_{n \in \mathcal{N}_1} \hat{y}_n & \nonumber \\
	\text{s.t.} \qquad & \sum_{n \in \mathcal{N}_0} \hat{y}_n \leq UB_u \label{constr:maxUB}\\
	& \hat{y}_n - \sum_{k \in \mathcal{\tilde{K}}} z_{nk}q_k \leq 0 & \forall n \in \mathcal{N}_1 \nonumber \\
	& \tilde{K}\hat{y}_n - \sum_{k \in \mathcal{\tilde{K}}} z_{nk}q_k \geq 0 & \forall n \in \mathcal{N}_0 \nonumber \\
	& \sum_{k \in \mathcal{\tilde{K}}} q_k \leq K \nonumber \\
	& q_k, \hat{y}_n \in \{0,1\} \nonumber .
\end{align}
Here, $\hat{y}$ is as before, and $q_k$ is a binary variable that indicates whether AND clause $k$ is included in the final Boolean phrase in DNF. The parameter $z_{nk}$ is equal to one when sample $n$ satisfies AND clause $k \in \mathcal{\hat{K}}$, and zero otherwise. Note that as the AND clauses are pre-definded, this is a parameter, not a variable. The maximum number of AND clauses included in the final Boolean statement is limited to at most a predetermined number $K$ to control the complexity of the statement.

\emph{The sub routine} In each iteration the same sub routine is executed for a predefined set of upper bounds $\{UB_1,...,UB_U\}$ on constraint \eqref{constr:maxUB}. The sub routine begins by solving $(MP)_u$ using all the AND clauses that were generated so far, denoted by $\hat{\mathcal{K}}$. If the objective value of $(MP)_u$ is at most $\tau_u$, where $\tau_u$ is the predefined desired objective value of $(MP)_u$, the sub routine for $UB_u$ ends. If the objective value of $(MP)_u$ is above $\tau_u$ the sub problem $(SP)_u$ is solved. As before, solving the sub problem for all $n \in \mathcal{N}$ is computationally challenging. Therefore the sub problem is solved only for a subset of the samples. First the set of false negatives corresponding to the solution to $(MP)_u$, denoted as $\mathcal{F}$, is computed. These false negatives are the class 1 samples for which no AND clause exists yet, or for which no AND clause exists that, in combination with the other available AND clauses, yields a good solution to $(MP)_u$. If $|\mathcal{F}| > N_s$, a random subset $\hat{\mathcal{N}}_1$ of $\mathcal{F}$ is selected. If $|\mathcal{F}| < N_s$ the set $\hat{\mathcal{N}}_1$ is set equal to $\mathcal{F}$. $(SP)$ is then solved for all samples in $\mathcal{N}_0 \cup \hat{\mathcal{N}}_1$. This ensures that a new AND clause is created that has the potential to increase the number of true positives when added to the most recently created Boolean phrase. 
The resulting new AND clauses are added to $\hat{\mathcal{K}}$.

\emph{The master problem} Once an objective value below $\tau_u$ has been reached for all sub routine master problems $(MP)_u$, the master problem can be solved using the obtained pool of AND clauses $\mathcal{\hat{K}}$. The master problem selects those AND clauses that constitute the best Boolean phrase in DNF in terms of classification accuracy. Let $\mathcal{\hat{K}}$ be a pool of AND clauses. The master problem is formulated as follows:
\begin{align}
	(MP) \qquad \min_{q,\hat{y}} \qquad & \sum_{n \in \mathcal{N}_0} w_n\hat{y}_n + \sum_{n \in \mathcal{N}_1} w_n(1-\hat{y}_n) & \\
	\text{s.t.} \qquad & \hat{y}_n - \sum_{k \in \mathcal{\tilde{K}}} z_{nk}q_k \leq 0 & \forall n \in \mathcal{N}_1 \\
	& \tilde{K}\hat{y}_n - \sum_{k \in \mathcal{\tilde{K}}} z_{nk}q_k \geq 0 & \forall n \in \mathcal{N}_0 \\
	& \sum_{k \in \mathcal{\tilde{K}}} q_k \leq K \\
	& q_k, \hat{y}_n \in \{0,1\}.
\end{align}
Here, $w$, $\hat{y}$, $z_{nk}$ and $q_k$ are as before.

\emph{IRELAND} Solving the sub routine for various upper bounds on the number of false positives gives AND clauses that represent various trade-offs between sensitivity and specificity. This allows IRELAND to select those AND clauses from the pool that together yield the best balanced accuracy. Note that this approach is highly parallelizable: the sub routine is carried out for $U$ upper bounds, hence the algorithm can solve up to $U$ optimization problems in parallel depending on the number of available threads.

\subsection{Generating the sensitivity-specificity trade-off curve}\label{sec:meth:Pareto}
IRELAND creates a pool of AND clauses with various sets of true and false positives and negatives, from which the master problem selects those that together yield the best balanced accuracy. This pool can be used to efficiently generate the sensitivity-specificity trade-off curve by solving a slight adaptation of the master problem. Two adaptations are used: $(MP)_\text{sens}$ maximizes sensitivity while placing a lower bound on the specificity, while $(MP)_\text{spec}$ maximizes the specificity while placing a lower bound on the sensitivity. By varying the lower bounds the sensitivity-specificity trade-off curve can be obtained. As $(MP)_\text{sens}$ and $(MP)_\text{spec}$ only solve which AND clauses from the pool $\hat{\mathcal{K}}$ are used without generating new AND clauses, generating the trade-off curve can be done within a limited amount of time.

Initially $(MP)_\text{sens}$ and $(MP)_\text{spec}$ are solved for a lower bound equal to zero on the specificity respectively the sensitivity to obtain the extreme points on the trade-off curve. Then in every iteration the algorithm searches for two neighboring points on the trade-off curve for which the sensitivities or specificities differ by more than a predetermined threshold. When two neighboring points with a difference in sensitivity (specificity) larger than the threshold are found, $(MP)_\text{spec}$ ($(MP)_\text{sens}$) is solved with a lower bound on the sensitivity (specificity) that is equal to the average sensitivity (specificity) of the two identified solutions on the trade-off curve. This procedure is repeated until there are no gaps larger than the threshold that can still be improved upon.

\subsection{Datasets}\label{sec:meth:Datasets}
Datasets were generated for various numbers of samples $N$ and numbers of features $J$. For each dataset a random input matrix $X$ was generated, as well as random Boolean DNF statements for given number of clauses $K$ and maximum number of features per clause $M$. These Boolean DNF phrases were used to generate $y$ from $X$. The dataset was only retained if it had at least 25\% cases and at least 25\% controls, else a new dataset for the given $N$, $J$, $K$ and $M$ was generated. For some combinations of $N$, $J$, $K$ and $M$ no dataset with a proper case/control ratio was found after 25 attempts, so that combination of parameters was dropped. 

Two collections of datasets were generated. The first collection contains 128 datasets with no noise introduced. This means that the optimal Boolean phrase in DNF yields a classification error and Hamming loss of 0. This collection of datasets is referred to as the no noise collection. 
Additionally 118 datasets with noise were generated. These datasets were generated in the same way as the noiseless datasets, except that a pre-determined fraction of the labels is inverted, meaning that if the sample was a case it becomes a control and vice versa. The error rates used were 1\%, 2.5\% and 5\%.


\section{Experiments and results}\label{sec:Results}
In this section results on the following topics are presented. In Section \ref{sec:res:MILP} the six MILP formulations from Table \ref{tab:overviewModels} are compared based on solution time. Section \ref{sec:res:hyperparameters} discusses the hyperparameter optimization for IRELAND. The performances of the original MILP and IRELAND are compared based on objective value and runtime for datasets of various sizes with and without noise in sections \ref{sec:res:nonoise} and \ref{sec:res:noise}, respectively. In Section \ref{sec:res:dash} the performance of IRELAND is compared to the model proposed by \cite{dash2018boolean}, which is considered the current state-of-the-art. Results for generating the sensitivity-specificity trade-off curve are presented in Section \ref{sec:res:Pareto}.

\subsection{The formulation of constraints for AND clauses largely affects solution time of the original LP models}\label{sec:res:MILP}
The six model formulations summarized in Table \ref{tab:overviewModels} were tested on the no noise datasets, and the results were compared based on solution times. All models were solved using the Gurobi 9.0.2 optimizer (Gurobi Optimization,
Inc., Houston, USA) interfaced with Python version 3.7.7 on a computer with an Intel
i7-9700 processor. Gurobi used 4 threads and was stopped after 300 seconds. Note that a problem that is solved to optimality within 300 seconds finds the optimal objective value of 0.

\begin{figure}
    \centering
    \subfloat[width=0.45\textwidth][{All decision variables are binary.}]{\includegraphics[width=0.45\textwidth]{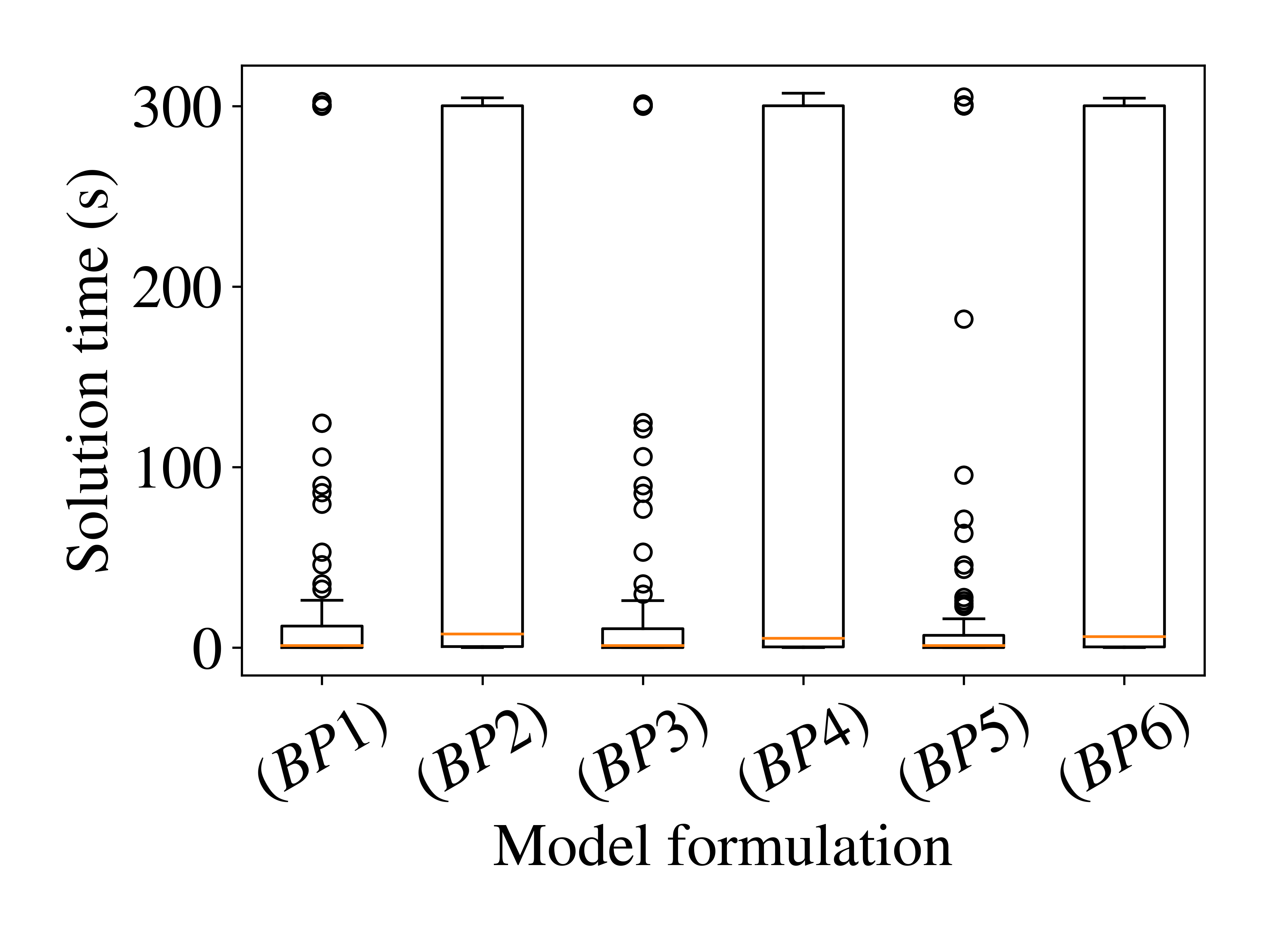}}
    \quad
    \subfloat[width=0.45\textwidth][Decision variables are relaxed whenever this does not alter the solution.]{\includegraphics[width=0.45\textwidth]{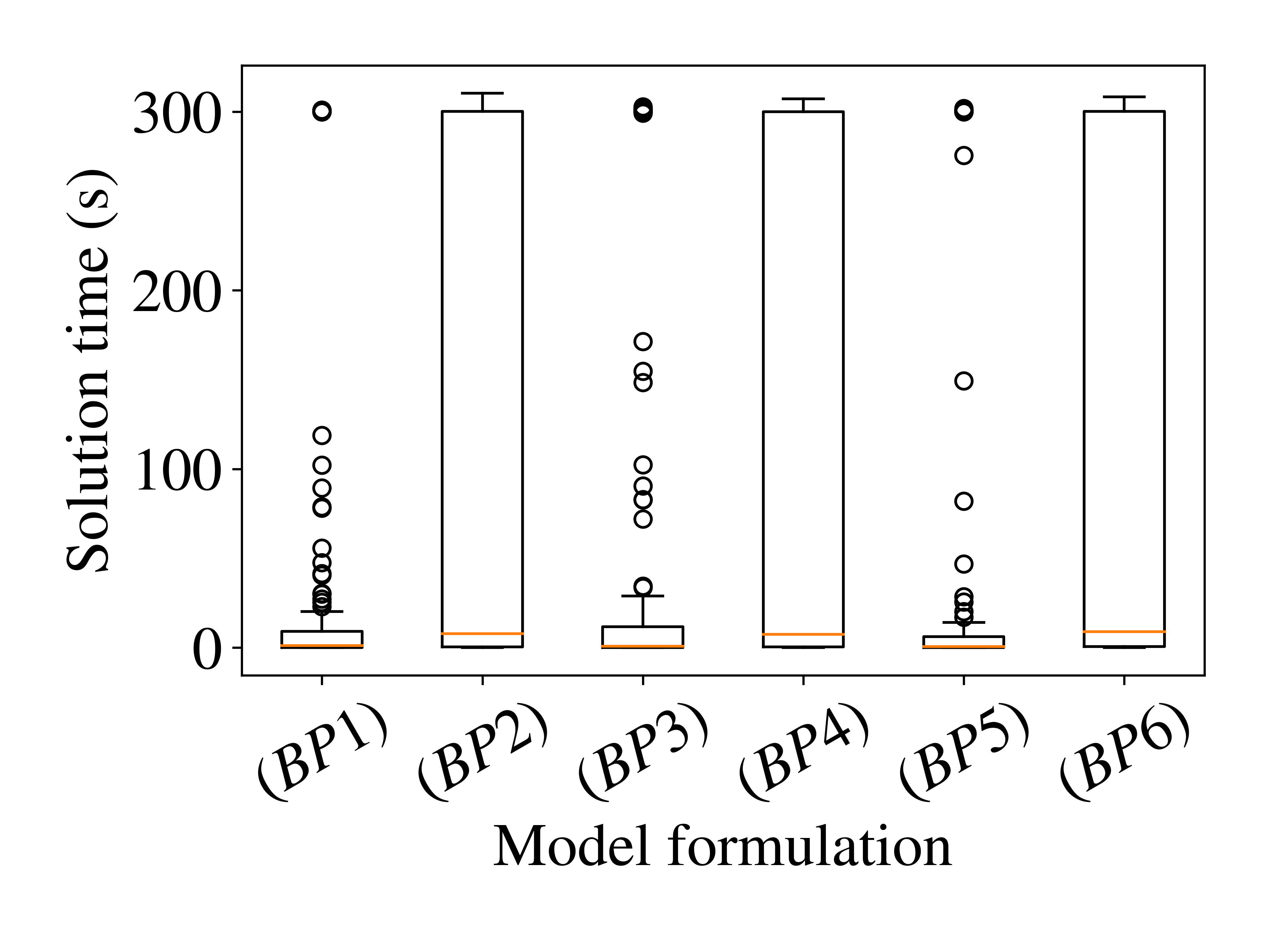}}
    \caption{Runtimes obtained with the six MILP formulations. The models were tested on the datasets in the no noise collection, where the optimal accuracy is known to be zero. The runtimes are limited to 300s.}
    \label{fig:baseMIPresults}
\end{figure}

The results are presented in Figure \ref{fig:baseMIPresults} for the models with all variables binary (Figure \ref{fig:baseMIPresults}a), as well as their LP relaxation where only those variables are relaxed that do not alter the optimal solution (Figure \ref{fig:baseMIPresults}b). The results show that the formulations that contain AND contraints \eqref{eq:AND1} largely outperform the formulations that contain constraints \eqref{eq:AND2}. The choice of OR constraints and objective function does not significantly influence the solution times. Additionally, a two-sided t-test was conducted to test the null hypothesis that the solution time of the formulation with all variables binary is equal to the solution time of that same formulation with some of the variables relaxed. From the results it can be concluded that there is no statistically significant difference in solution times between models with all variables binary and some relaxed variables ($\alpha=0.05$) except for $(BP2)$ ($p=0.0074$), where the relaxation is slightly slower (on average 3.1 seconds).

\subsection{Hyperparameter selection for IRELAND}\label{sec:res:hyperparameters}
\begin{figure}
    \centering
    \includegraphics[width=\textwidth]{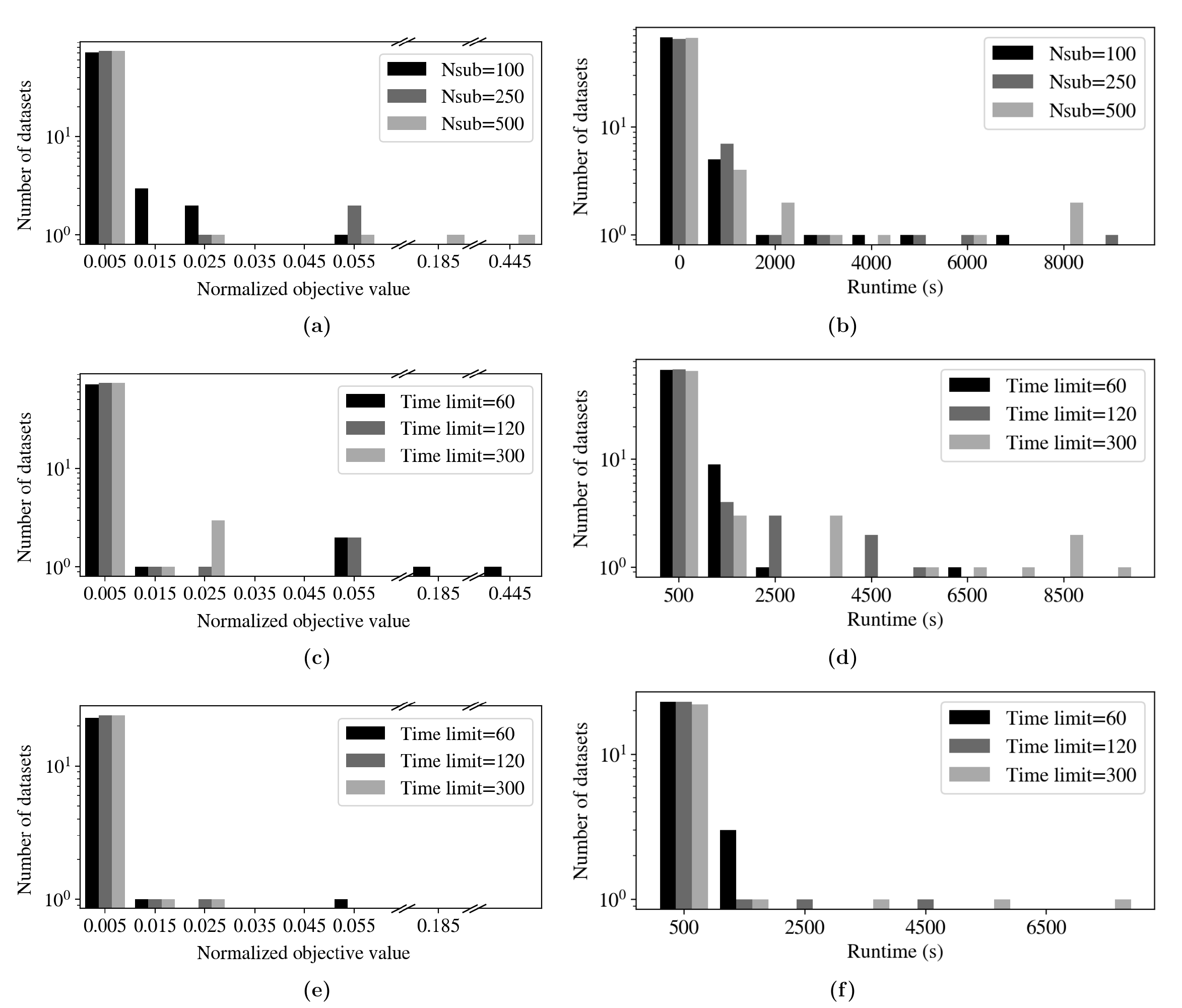}
    \caption{Histograms showing the objective function values (a and c) and runtimes (b and d) obtained when solving for noiseless datasets with IRELAND using various values for $N_s$ (a and b) and the time limit for the master- and sub problem (c and d). In figures a and b results are shown for all noiseless datasets and all choices for a time limit of 30, 120 and 300 seconds. In figures c and d results are shown for all noiseless datasets and $N_s=100$.}
    \label{fig:ParameterSelection}
\end{figure}

The no noise datasets were run for $UB=0$ only on a pc with four threads. 
To identify the optimal choice for $N_s$ and the time limit for each solve of the master- and sub problem, IRELAND was tested on 26 noiseless datasets for $N_s=100, 250$ and $500$, and for a time limit of 60, 120 and 300 seconds. Histograms of the objective values and total runtimes per choice of $N_s$ aggregated over the 26 noiseless datasets and all three time limits are shown in Figures \ref{fig:ParameterSelection}a and \ref{fig:ParameterSelection}b, respectively. These histograms show that $N_s=100$ in general yields the lowest objective value and runtime. Similar histograms showing the objective values and runtimes per choice of the time limit aggregated over the 26 noiseless datasets and all choices for $N_s$ are shown in Figures \ref{fig:ParameterSelection}c and \ref{fig:ParameterSelection}d, respectively. When looking at the objective function values there are two outliers, both corresponding to a run with $N_s=500$. In order to choose a time limit that performs best given that $N_s=100$, Figures \ref{fig:ParameterSelection}e and \ref{fig:ParameterSelection}f present the same results but only for $N_s=100$. The histograms show that a time limit of 60 seconds yields high objective values and runtimes and is therefore unsuitable. Time limits of 120 and 300 seconds yield similar objective values, while a time limit of 300 seconds results in a much larger runtime than a time limit of 120 seconds. In the remainder of this work we therefore use $N_s=100$ and a time limit of 120 seconds.

\subsection{Performance of $(BP1)$ versus IRELAND on data without noise}\label{sec:res:nonoise}
\begin{figure}
     \centering
     \subfloat[width=0.45\textwidth][]{\includegraphics[width=0.45\textwidth]{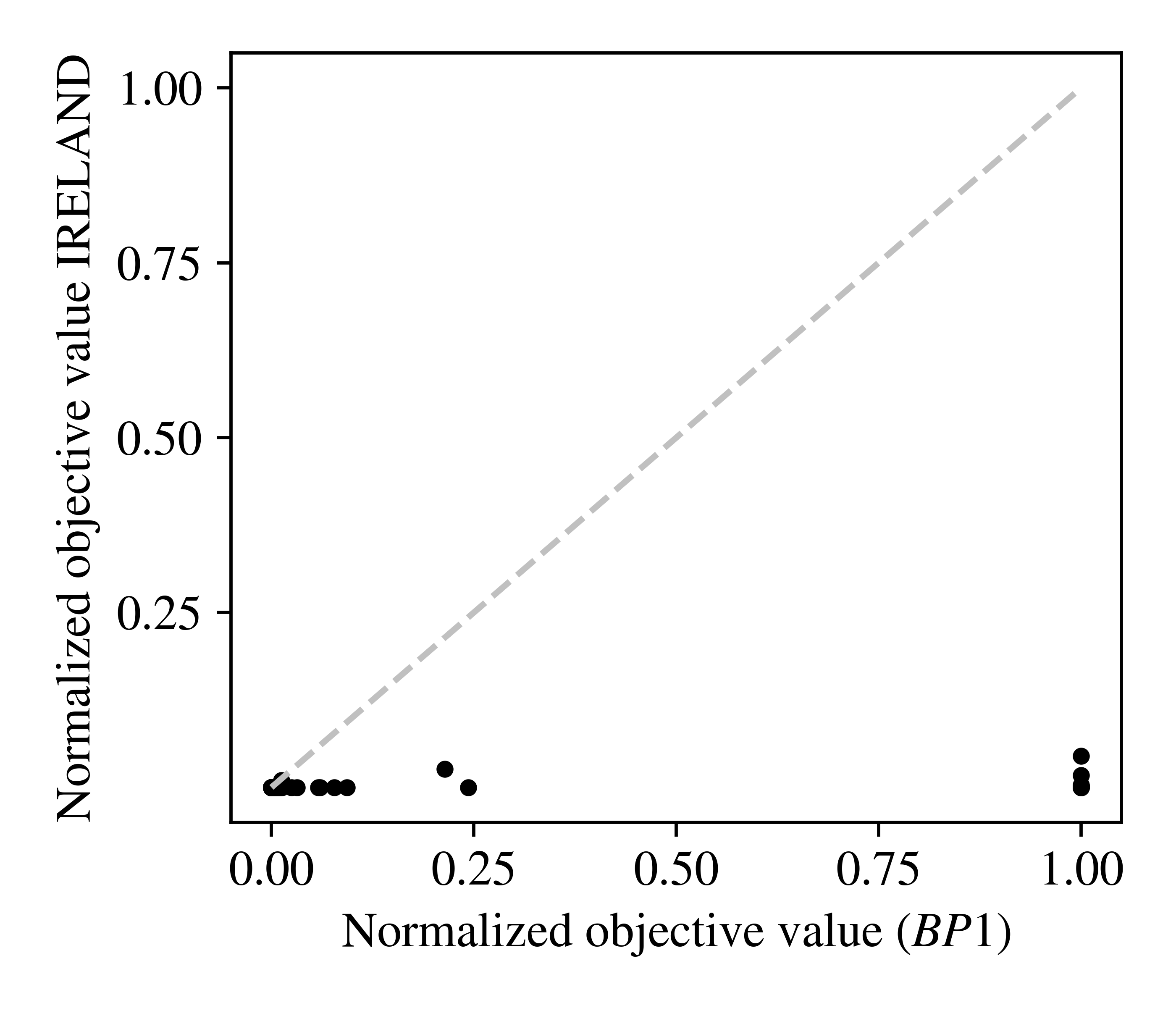}}
     \subfloat[width=0.45\textwidth][]{\includegraphics[width=0.45\textwidth]{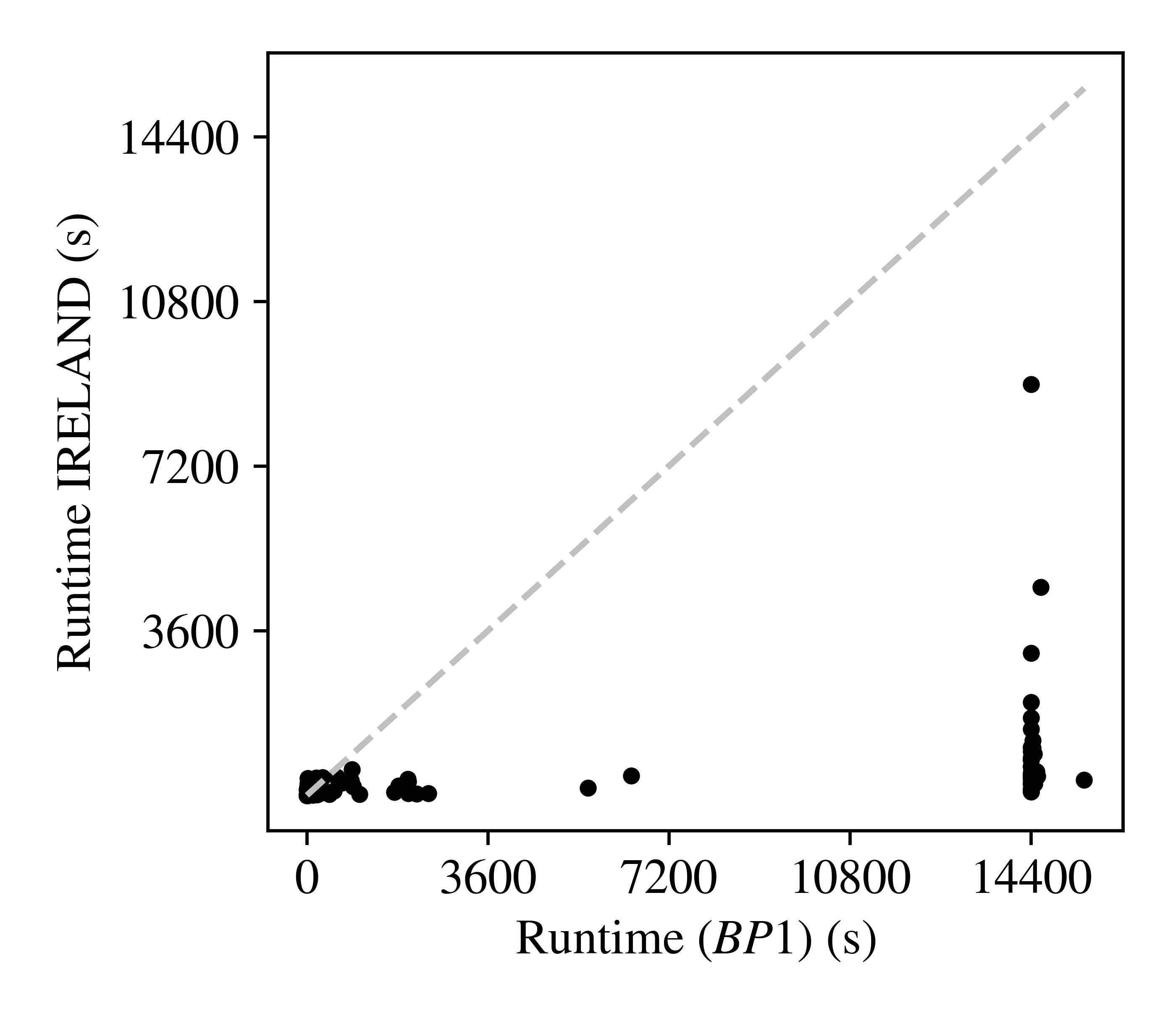}}
     \caption{Comparison of the performance of $(BP1)$ versus IRELAND on datasets without noise in terms of normalized objective function value (a) and runtime in seconds (b). Each dot represents a dataset, for which the normalized objective value and the runtime of $(BP1)$ are shown on the horizontal axis, and the normalized objective value and runtime of IRELAND are shown on the vertical axis. The dashed line indicates equal performance between the methods.}
     \label{fig:results_nonoise}
\end{figure}

$(BP1)$ and IRELAND were both used to solve the classification problem of the 128 instances in the data collection without noise for a maximum runtime of four hours. Performances were compared based on objective values as well as runtimes as shown in Figure \ref{fig:results_nonoise}. Each dot represents a dataset for which the normalized objective function values, that is, the objective function values divided by the number of samples, (Figure \ref{fig:results_nonoise}a) and runtimes (Figure \ref{fig:results_nonoise}b) obtained with $(BP1)$ are shown on the horizontal axis, and those for IRELAND on the vertical axis. A diagonal line is included to indicate equal objective function values and runtimes for the two solution methods. Note that for these datasets a solution with an objective value equal to zero exists, as they do not contain any noise. 

Figure \ref{fig:results_nonoise}a shows that for the majority of datasets both $(BP1)$ and IRELAND find a near-optimal solution. $(BP1)$ found the optimal solution for 90 datasets, and IRELAND found the optimal solution for 119 datasets. For all datasets where IRELAND did not find the optimal solution, the obtained objective function value was the same as or lower than the objective function value obtained with $(BP1)$. For ten datasets $(BP1)$ ran out of memory, hence an objective value of 1 and the maximum runtime of 14,400 seconds (four hours) were assigned to these datasets. For these datasets IRELAND did find solutions with a normalized objective function value between 0.0 and 0.045.

The results in Figure \ref{fig:results_nonoise}b show that for the datasets where $(BP1)$ has runtimes below approximately 90 seconds, IRELAND cannot improve upon this. For those datasets where $(BP1)$ finds an optimal solution within four hours, IRELAND finishes within 20 minutes. The datasets that could not be solved by $(BP1)$ due to memory issues or a time limit of four hours were all solved by IRELAND. In most cases IRELAND finished within an hour, only for two of datasets it needed 1 hour 15 minutes and 2 hours 30 minutes, respectively.

\begin{figure}
    \centering
    \includegraphics[width=\textwidth]{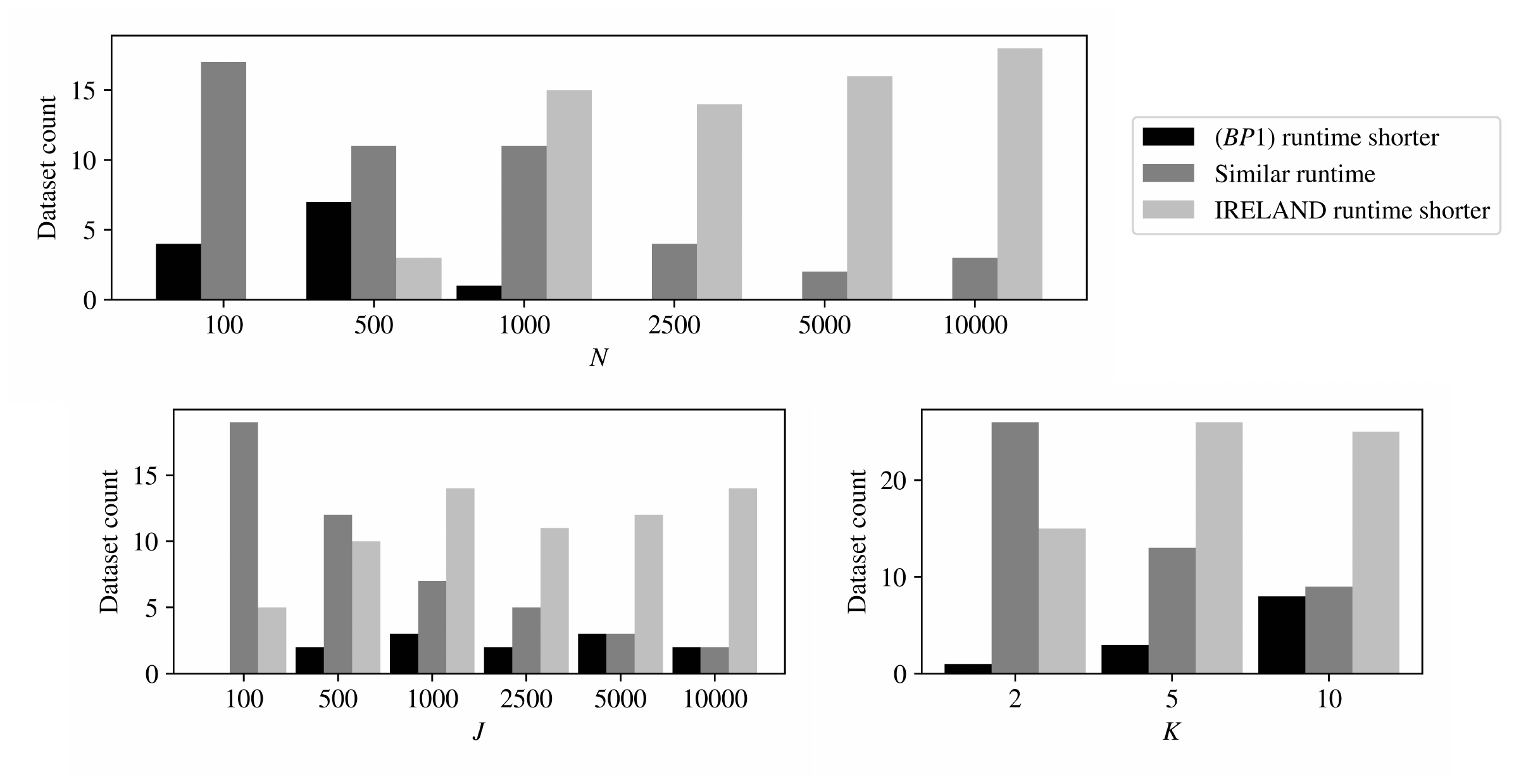}
    \caption{Histograms of the number of noiseless datasets for which $(BP1)$ has a better runtime than IRELAND (black), the runtimes do not differ by more than 30 seconds (dark gray) and IRELAND has a better runtime than $(BP1)$ (light gray), split per $N$, $J$ and $K$.}
    \label{fig:better_runtimes_perNKP}
\end{figure}

\begin{figure}
    \centering
    \includegraphics[width=0.8\textwidth]{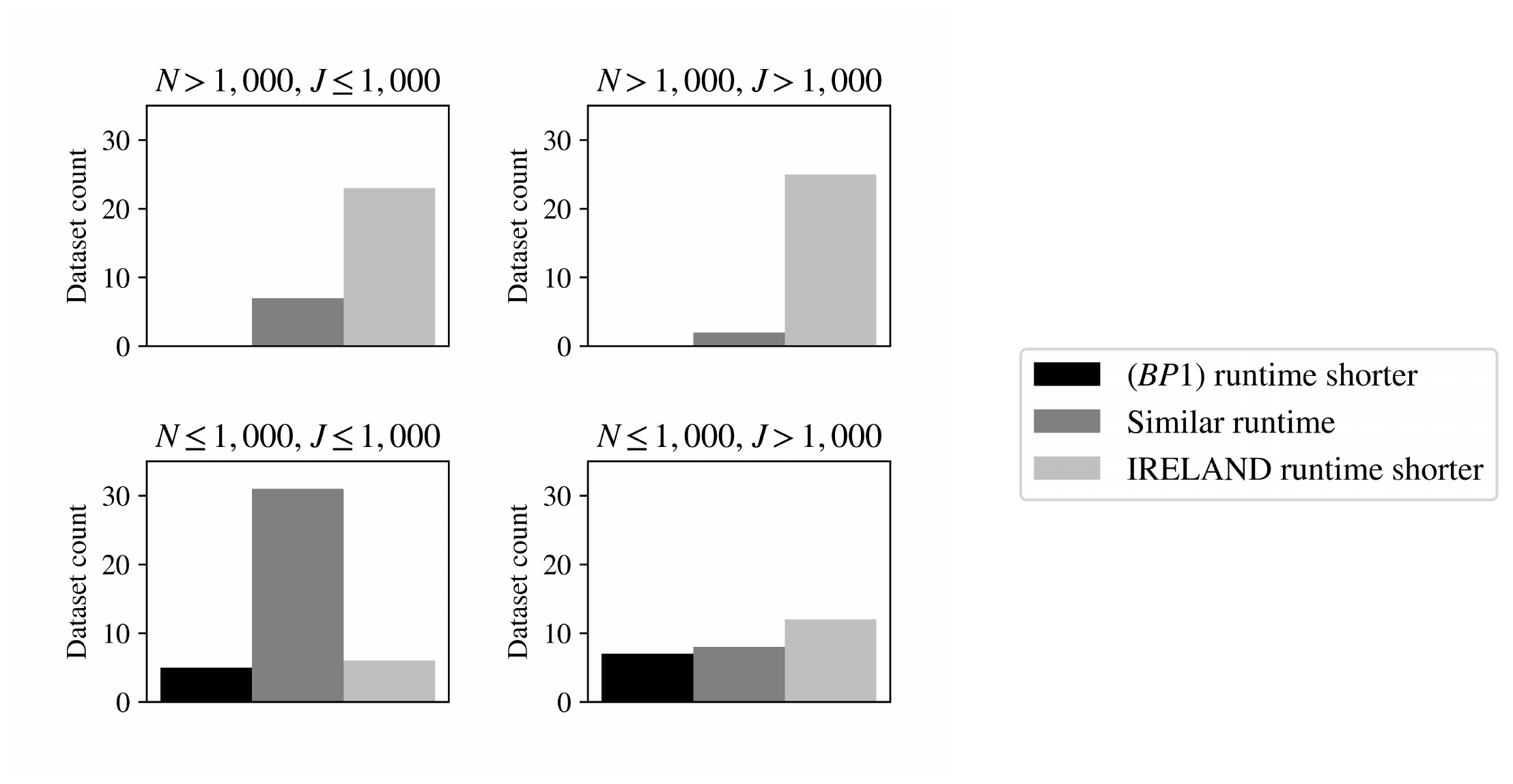}
    \caption{Histograms of the number of noiseless datasets for which $(BP1)$ has a better runtime than IRELAND (black), the runtimes do not differ by more than 30 seconds (dark gray) and IRELAND has a better runtime than $(BP1)$ (light gray), split by $N \leq 1,000$ versus $N > 1,000$, and by $J\leq 1,000$ versus $J>1,000$.}
    \label{fig:better_runtimes_splitNP}
\end{figure}

In order to see for which datasets it is best to use $(BP1)$ and for which to use IRELAND, Figure \ref{fig:better_runtimes_perNKP} shows the number of datasets for which $(BP1)$ has a lower runtime than IRELAND, the number of datasets for which IRELAND has a lower runtime than $(BP1)$ and the number of datasets for which the difference in runtime is less than 30 seconds, split by number of samples $N$, number of features $J$ and number of AND clauses $K$. For $N\leq 500$, $(BP1)$ has lower runtimes for most datasets, while for $N\geq 1,000$ IRELAND has a clear advantage over $(BP1)$. Figure \ref{fig:better_runtimes_perNKP} shows that when $J$ is large, there are datasets for which $(BP1)$ outperforms IRELAND. However, this is only the case when $N$ is at most 1,000, see Figure \ref{fig:better_runtimes_splitNP}. $K$ seems to be a weak indicator of which method performs best.

\subsection{Performance $(BP1)$ versus IRELAND on data with noise}\label{sec:res:noise}
The datasets with noise were run on a computer with 24 threads. The sub- and master problems were solved for six values of $UB$ in parallel, $UB_u \in (0.005,0.01,0.02,0.03,0.04,0.05)$, allowing Gurobi to use four threads for each optimization.

\begin{figure}
     \centering
     \subfloat[width=0.45\textwidth][]{\includegraphics[width=0.45\textwidth]{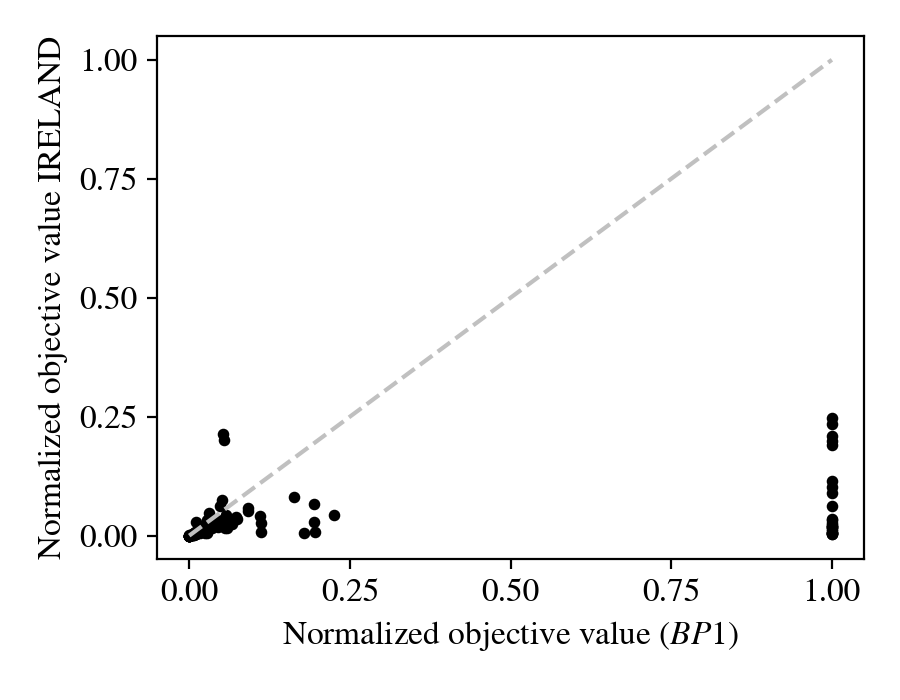}}
     \subfloat[width=0.45\textwidth][]{\includegraphics[width=0.45\textwidth]{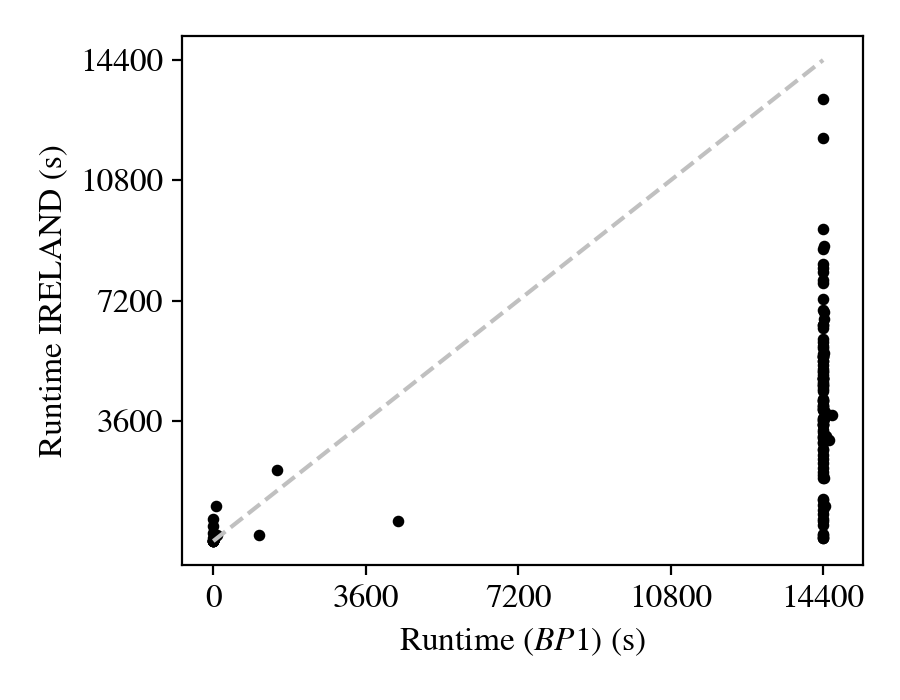}}
     \caption{Comparison of the performance of $(BP1)$ versus IRELAND on datasets with noise in terms of normalized objective function value (a) and runtime (b) in seconds. Each dot represents a dataset, for which the normalized objective value and the runtime of $(BP1)$ are shown on the horizontal axis, and the normalized objective value and runtime of IRELAND are shown on the vertical axis. The dashed line indicates equal performance between the methods.}
     \label{fig:results_noise}
\end{figure}

Figure \ref{fig:results_noise} shows a comparison between $(BP1)$ and IRELAND in terms of objective value (a) and runtime (b) for noisy datasets. Figure \ref{fig:results_noise}b shows that for 74 out of the 118 datasets $(BP1)$ required more than 4 hours of runtime. For another 25 datasets no solution was found at all as the system ran out of memory, hence these datasets were assigned an objective value of 1.0 and a runtime of 4 hours. For those datasets where $(BP1)$ did find a solution within the set time limit, Figure \ref{fig:results_noise} shows that for most datasets IRELAND outperformed $(BP1)$.

\begin{figure}
    \centering
    \includegraphics[width=\textwidth]{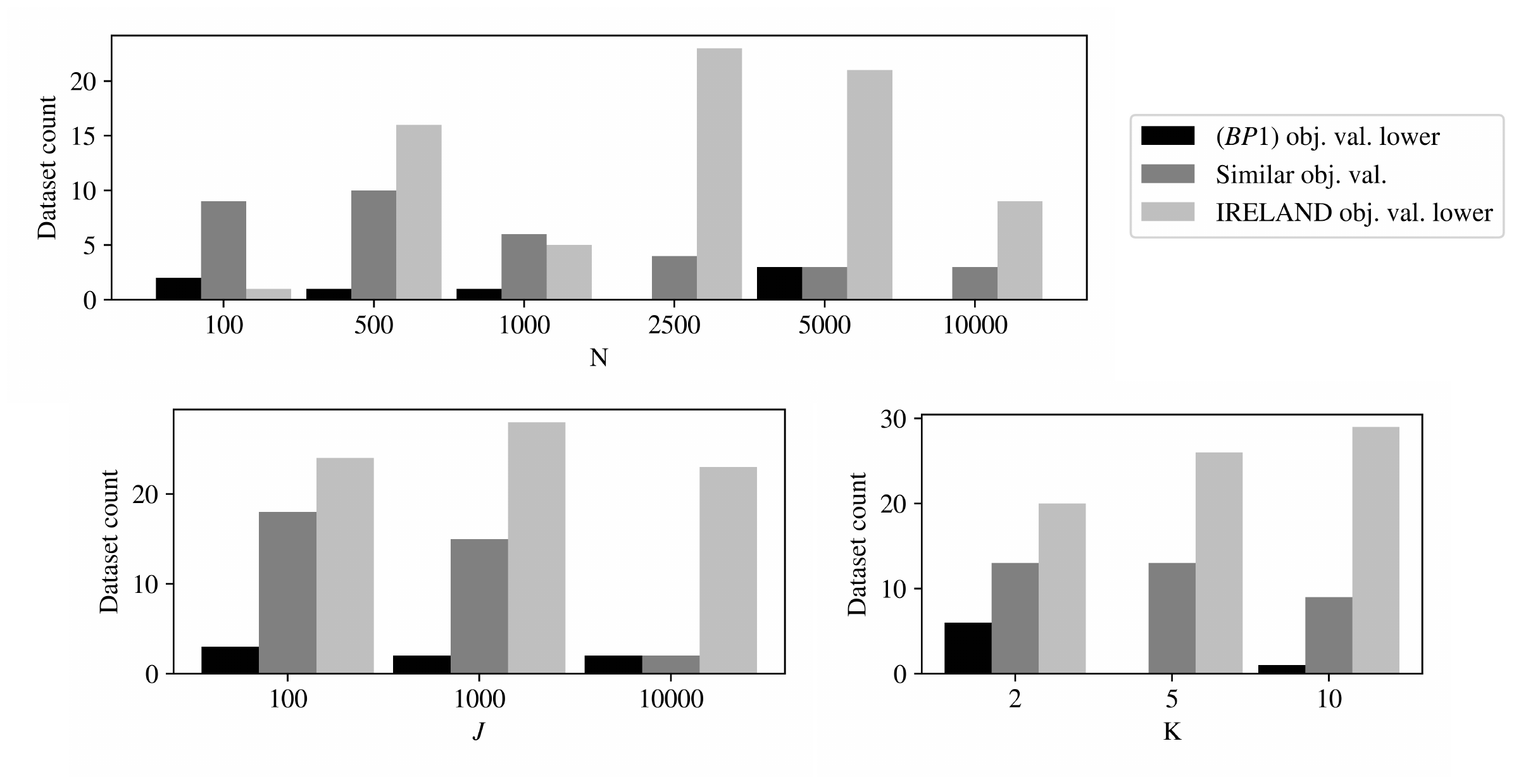}
    \caption{Histograms of the number of datasets (with noise) for which $(BP1)$ has a better objective value than IRELAND (black), the objective values do not differ by more than 0.005 (dark gray) and IRELAND has a better objective function value than $(BP1)$ (light gray), split per $N$, $J$ and $K$.}
    \label{fig:better_objvals_perNKP_withErrors}
\end{figure}

\begin{figure}
    \centering
    \includegraphics[width=\textwidth]{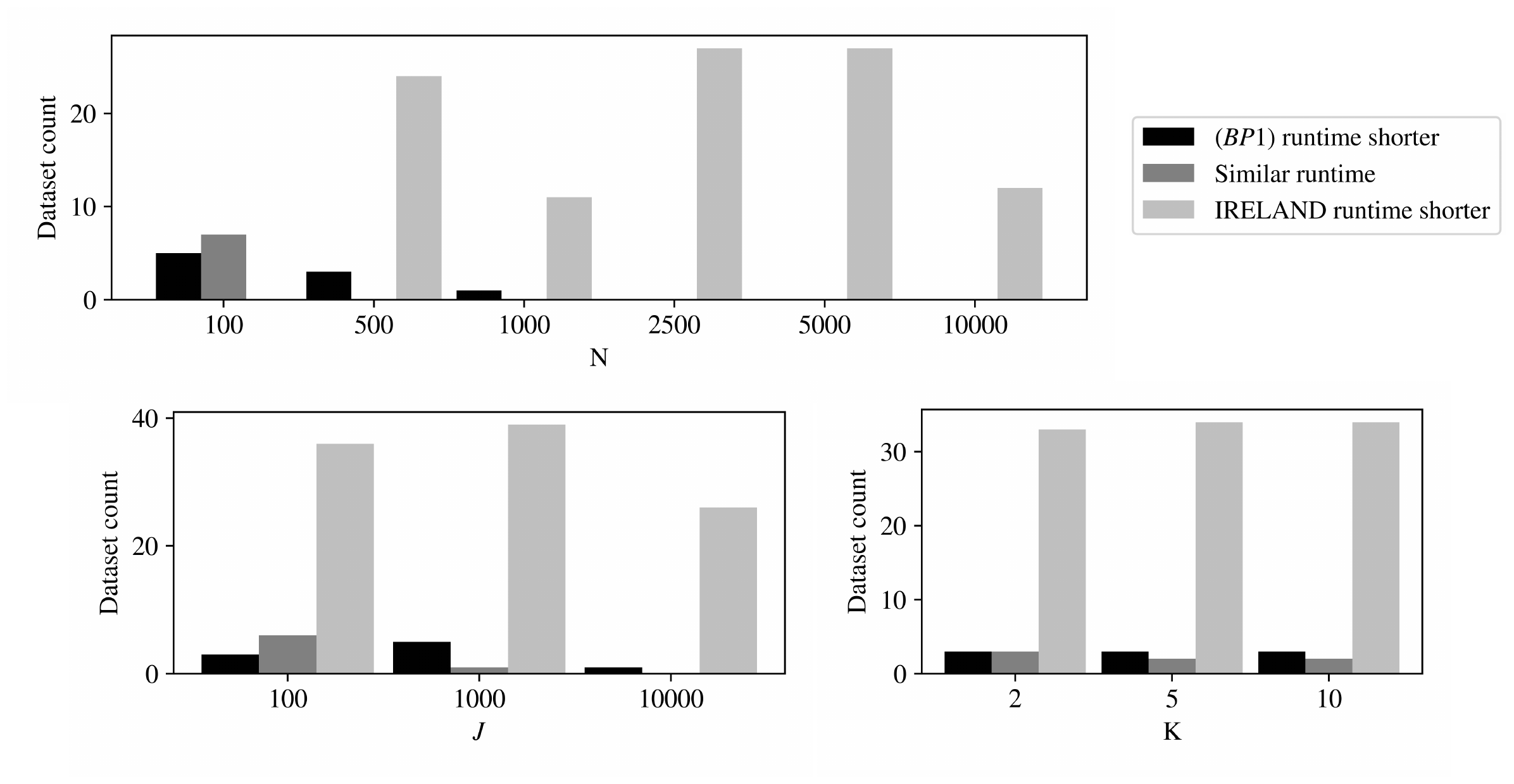}
    \caption{Histograms of the number of datasets (with noise) for which $(BP1)$ has a better runtime than IRELAND (black), the runtimes do not differ by more than 30 seconds (dark gray) and IRELAND has a better runtime than $(BP1)$ (light gray), split per $N$, $J$ and $K$.}
    \label{fig:better_runtimes_perNKP_withErrors}
\end{figure}

\begin{figure}
    \centering
    \includegraphics[width=\textwidth]{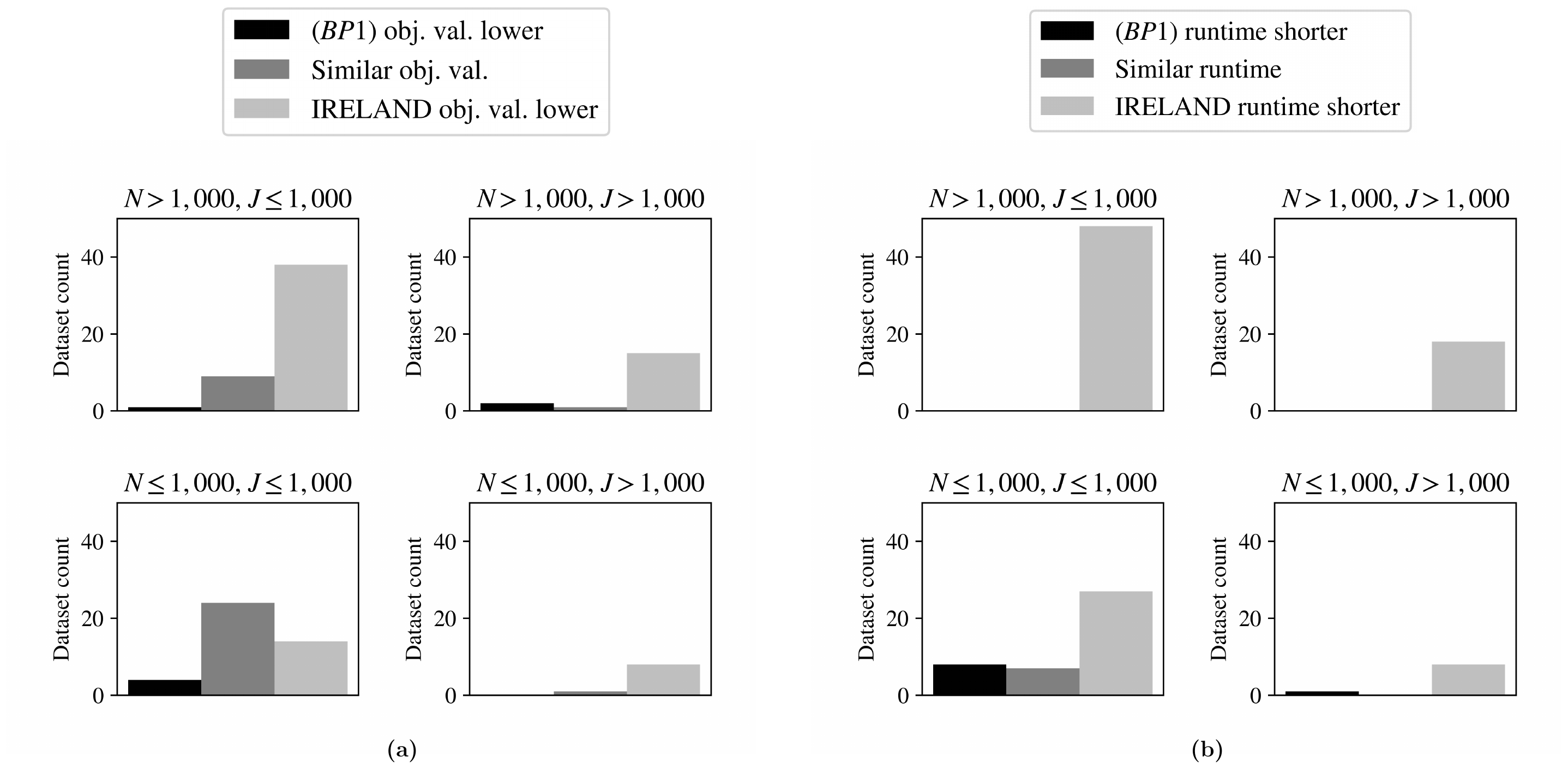}
    \caption{Histograms of the number of datasets (with noise) for which $(BP1)$ has a better objective value (a) or runtime (b) than IRELAND (black), the objective values (a) or runtimes (b) do not differ by more than 30 seconds (dark gray) and IRELAND has a better objective value (a) or runtime (b) than $(BP1)$ (light gray), split by $N < 1,000$ versus $N\geq1,000$, and by $P\leq 1,000$ versus $J>1,000$.}
    \label{fig:better_runtimes_splitNP_withErrors}
\end{figure}

Figures \ref{fig:better_objvals_perNKP_withErrors} and \ref{fig:better_runtimes_perNKP_withErrors} show histograms of the number of datasets for which IRELAND outperformed $(BP1)$, $(BP1)$ outperformed IRELAND, or performance was similar in terms of objective and runtime, respectively, separated by number of samples $N$, number of features $J$ and number of AND clauses $K$. Similar to the noiseless setting, the histograms show that for $N \leq 500$ IRELAND often, but not always, ourperforms $(BP1)$ in terms of objective values and runtimes, while for $N\geq 1000$, the main purpose of this work, IRELAND always outperforms $(BP1)$. Figures \ref{fig:better_objvals_perNKP_withErrors} and \ref{fig:better_runtimes_perNKP_withErrors} seem to indicate that IRELAND outperforms $(BP1)$ when $J$ is large, while Figure \ref{fig:better_runtimes_splitNP_withErrors} shows that $N$ remains the most important indicator for when to choose IRELAND over $(BP1)$.

\subsection{Comparing IRELAND and BRCG}\label{sec:res:dash}
\begin{figure}
     \centering
     \subfloat[width=0.45\textwidth][]{\includegraphics[width=0.45\textwidth]{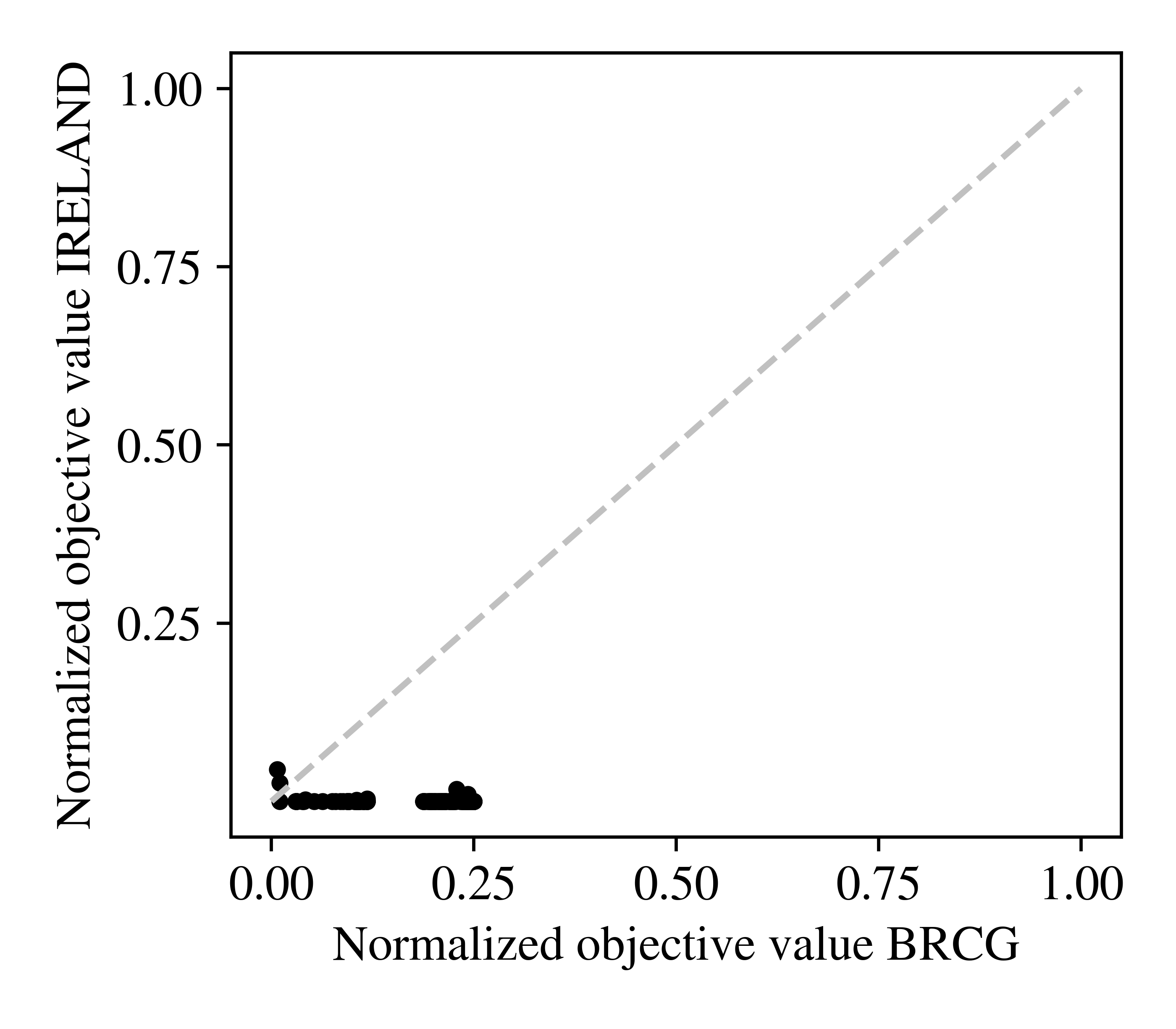}}
     \subfloat[width=0.45\textwidth][]{\includegraphics[width=0.45\textwidth]{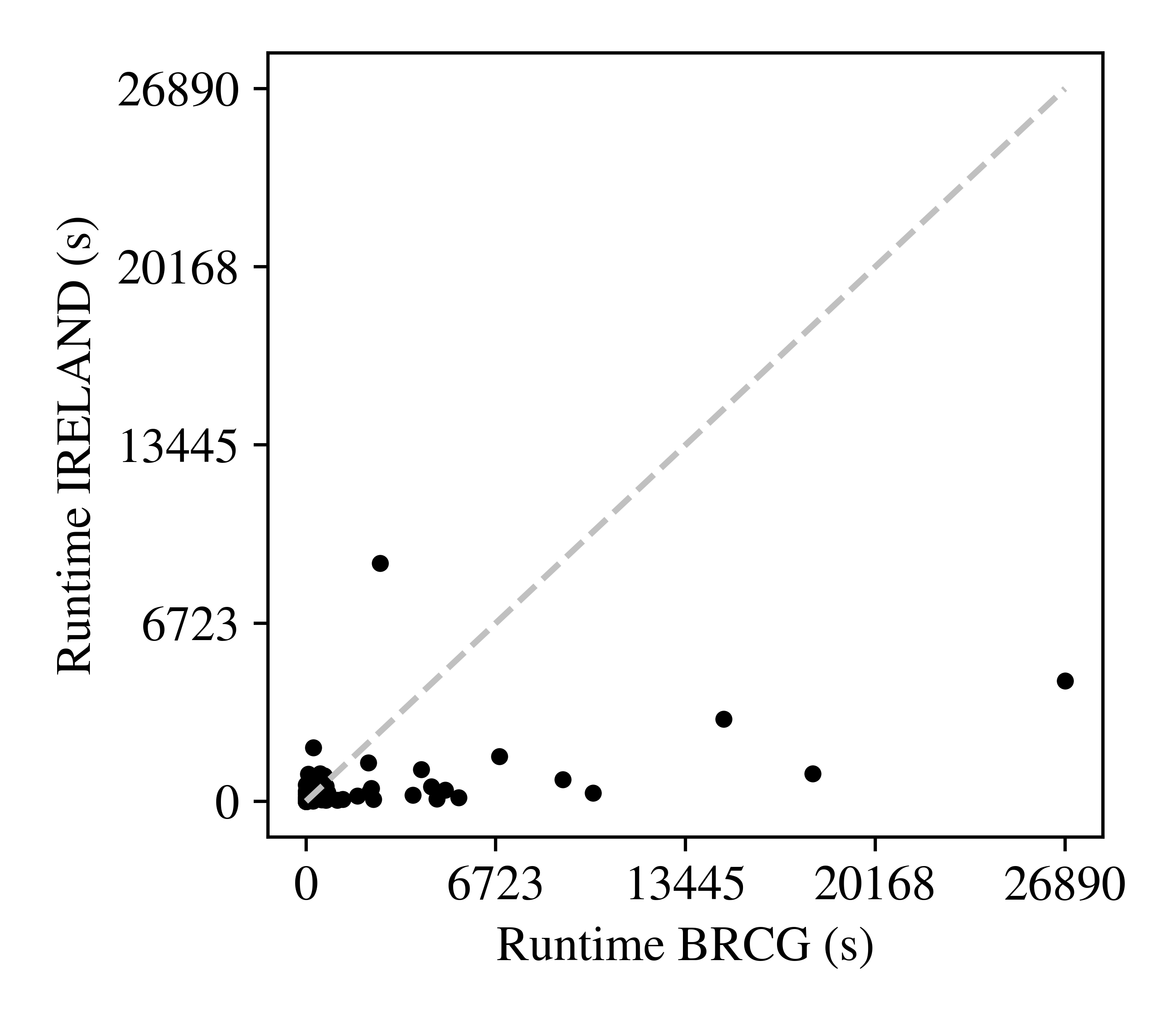}}
     \caption{Comparison of the performance of BRCG versus IRELAND on datasets without noise in terms of normalized objective function value (a) and runtime in seconds (b). Each dot represents a dataset, for which the normalized objective value and the runtime of BRCG are shown on the horizontal axis, and the normalized objective value and runtime of IRELAND are shown on the vertical axis. The dashed line indicates equal performance between the methods.}
     \label{fig:results_nonoise_dash}
\end{figure}

\begin{figure}
     \centering
     \subfloat[width=0.45\textwidth][]{\includegraphics[width=0.45\textwidth]{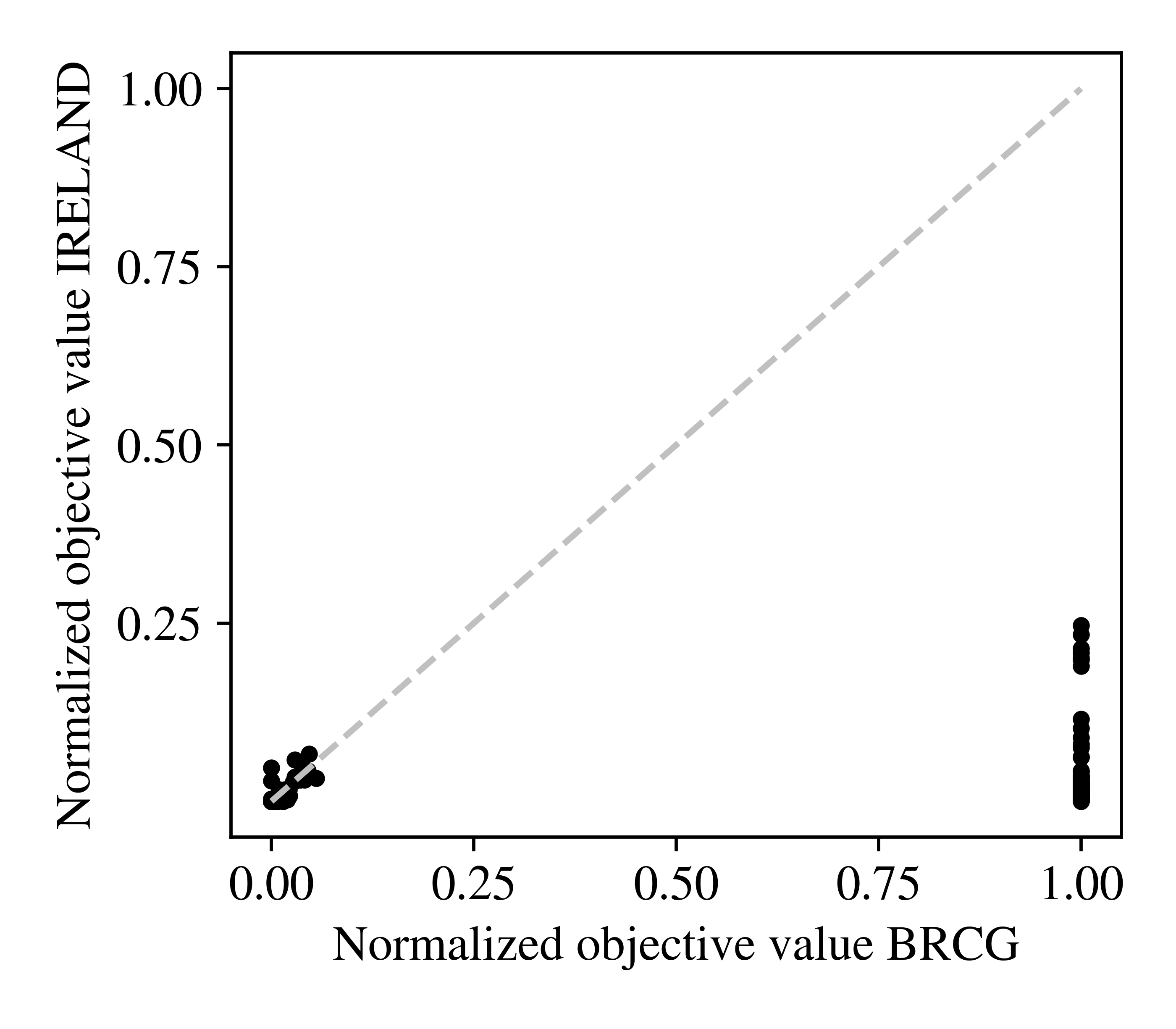}}
     \subfloat[width=0.45\textwidth][]{\includegraphics[width=0.45\textwidth]{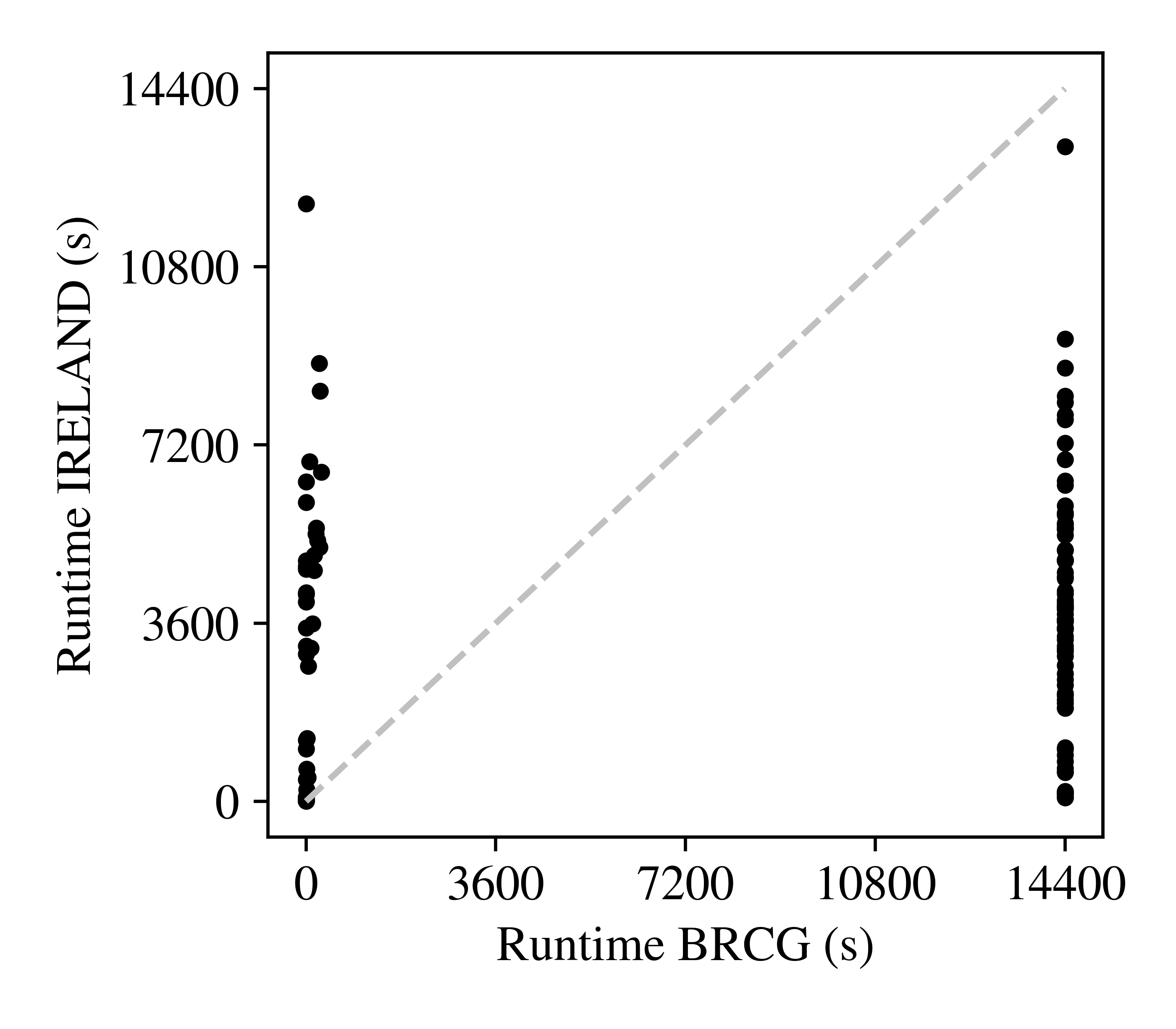}}
     \caption{Comparison of the performance of BRCG versus IRELAND on datasets with noise in terms of normalized objective function value (a) and runtime in seconds (b). Each dot represents a dataset, for which the normalized objective value and the runtime of BRCG are shown on the horizontal axis, and the normalized objective value and runtime of IRELAND are shown on the vertical axis. The dashed line indicates equal performance between the methods.}
     \label{fig:results_noise_dash}
\end{figure}

\begin{figure}
    \centering
    \includegraphics[width=\textwidth]{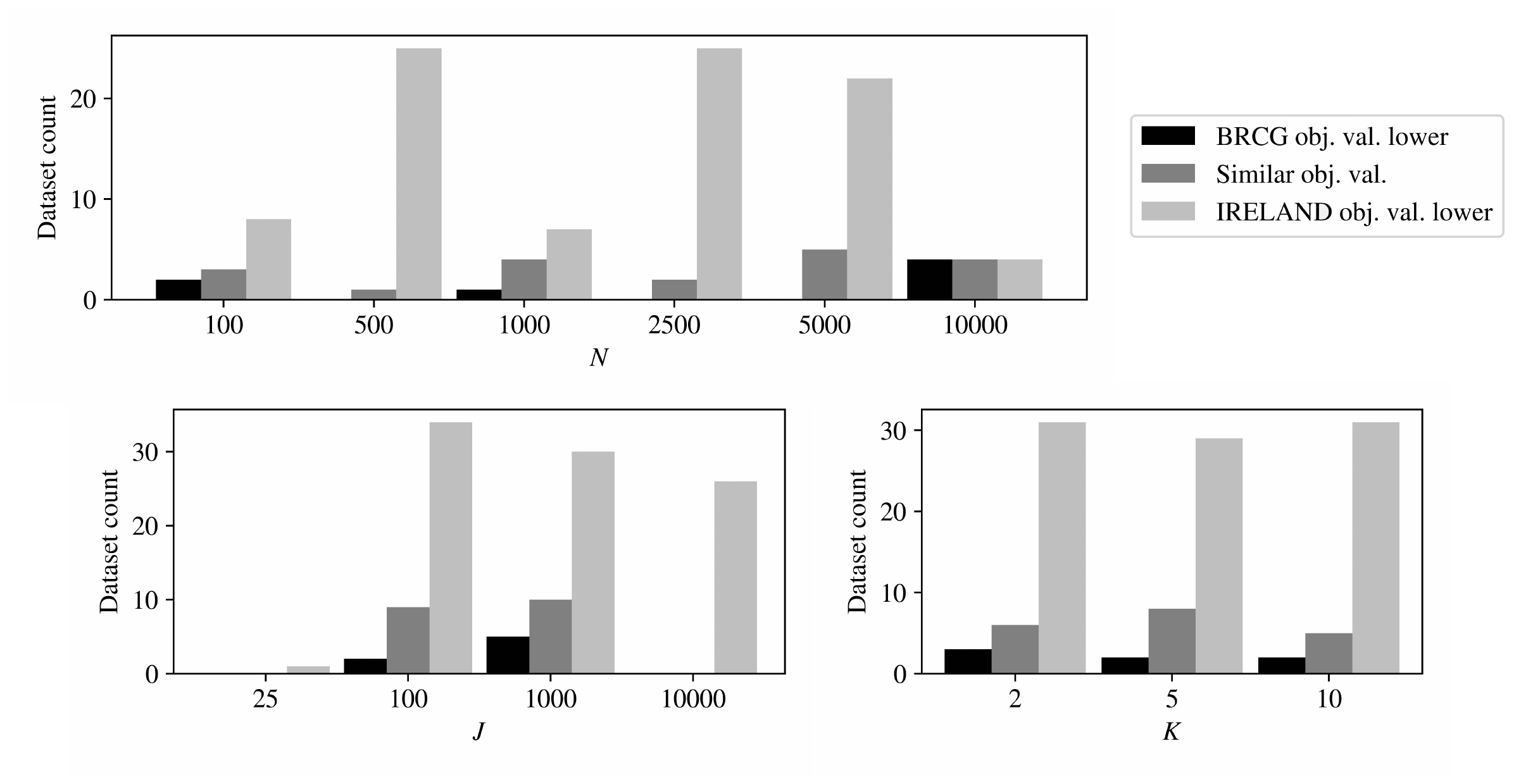}
    \caption{Histograms of the number of datasets (with noise) for which BRCG yields a better objective value than IRELAND (black), the objective values do not differ by more than 0.005 (dark gray) and IRELAND has a yields a better objective value than BRCG (light gray), split per $N$, $J$ and $K$.}
    \label{fig:better_setgen_Dash_perNKP_withErrors_obj}
\end{figure}

\begin{figure}
    \centering
    \includegraphics[width=\textwidth]{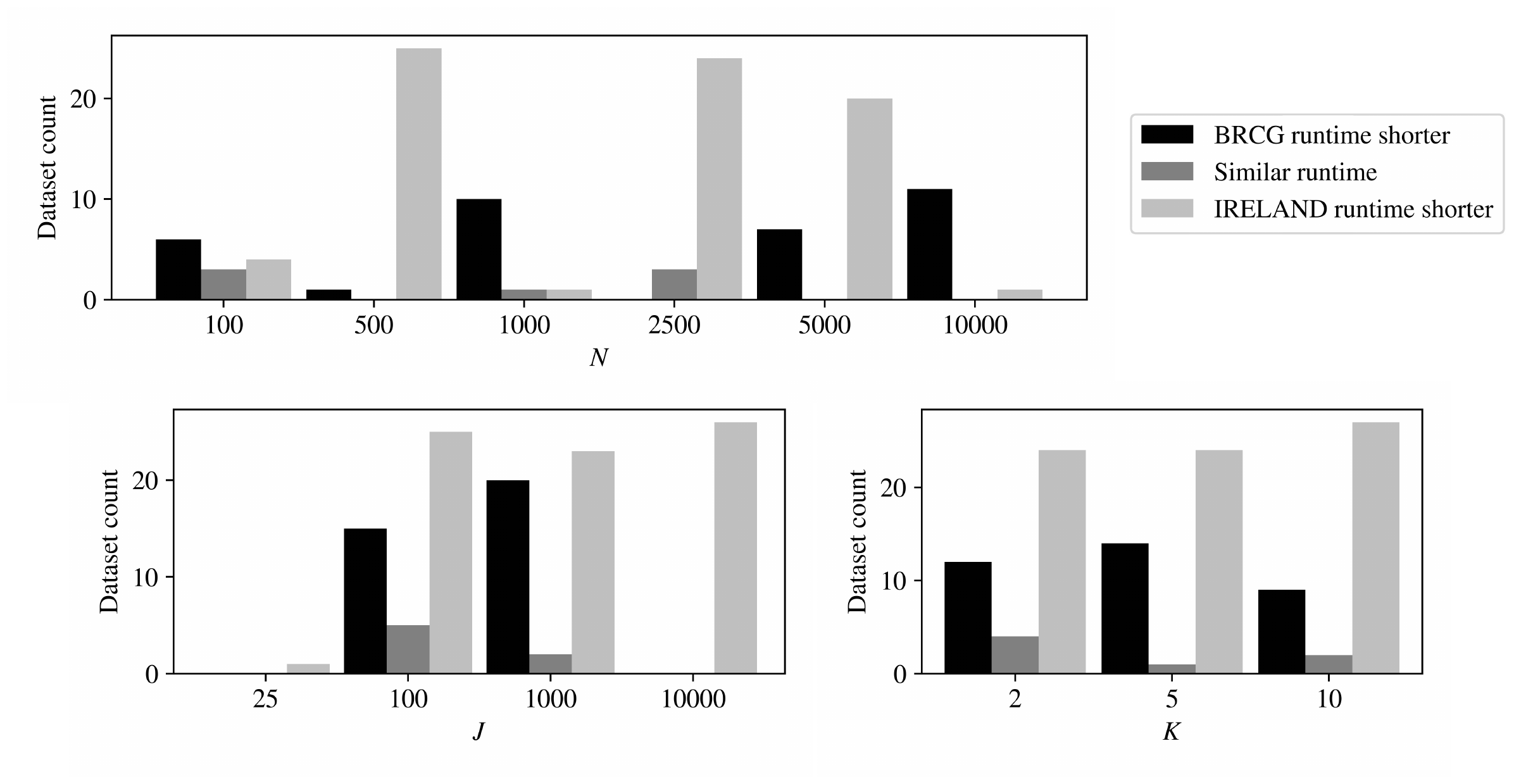}
    \caption{Histograms of the number of datasets (with noise) for which BRCG has a better runtime than IRELAND (black), the runtimes do not differ by more than 30 seconds (dark gray) and IRELAND has a better runtime than BRCG (light gray), split per $N$, $J$ and $K$.}
    \label{fig:better_setgen_Dash_perNKP_withErrors_runtime}
\end{figure}

\begin{figure}
    \centering
    \includegraphics[width=\textwidth]{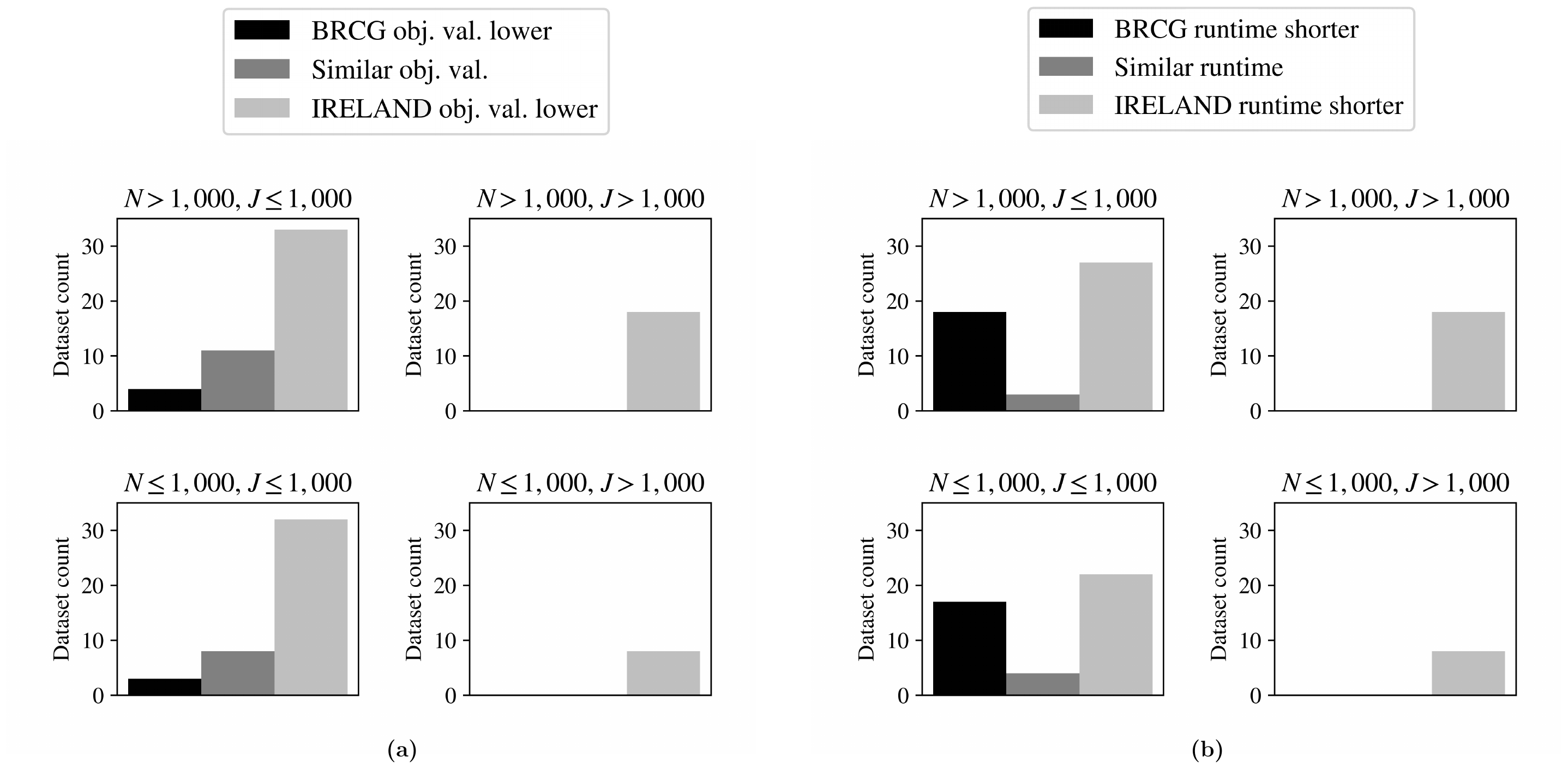}
    \caption{Histograms of the number of datasets (with noise) for which BRCG has a better objective value (a) or runtime (b) than IRELAND (black), the objective values (a) or runtimes (b) do not differ by more than 30 seconds (dark gray) and IRELAND has a better objective value (a) or runtime (b) than BRCG (light gray), split by $N < 1,000$ versus $N\geq1,000$, and by $P\leq 1,000$ versus $J>1,000$.}
    \label{fig:better_setgen_Dash_splitNP_withError}
\end{figure}

Recently \cite{dash2018boolean} implemented a column generation approach to the problem of generating Boolean phrases in DNF from binary data. The authors showed that their method, referred to as Boolean Rule Column Generation (BRCG), outperforms various state-of-the art approaches. Figures \ref{fig:results_nonoise_dash} and \ref{fig:results_noise_dash} compare the performances of BRCG and IRELAND for datasets with and without noise, respectively. For datasets without noise IRELAND outperforms BRCG in terms of both objective value and runtime for nearly all datasets. When noise is introduced we need to distinguish between two groups of datasets. For one group IRELAND and BRCG perform similarly in terms of objective function value, but IRELAND may require much more time than BRCG. For the second group BRCG cannot find a solution within four hours or runs out of memory, while IRELAND is able to find such a solution, often with a low objective value. Figures \ref{fig:better_setgen_Dash_perNKP_withErrors_obj}, \ref{fig:better_setgen_Dash_perNKP_withErrors_runtime} and \ref{fig:better_setgen_Dash_splitNP_withError} show that  IRELAND outperforms BRCG for datasets with a large number of features $J$.

\subsection{The sensitivity-specificity trade-off curve}\label{sec:res:Pareto}
\begin{figure}
    \centering
    \subfloat[]{\includegraphics[width=0.5\textwidth]{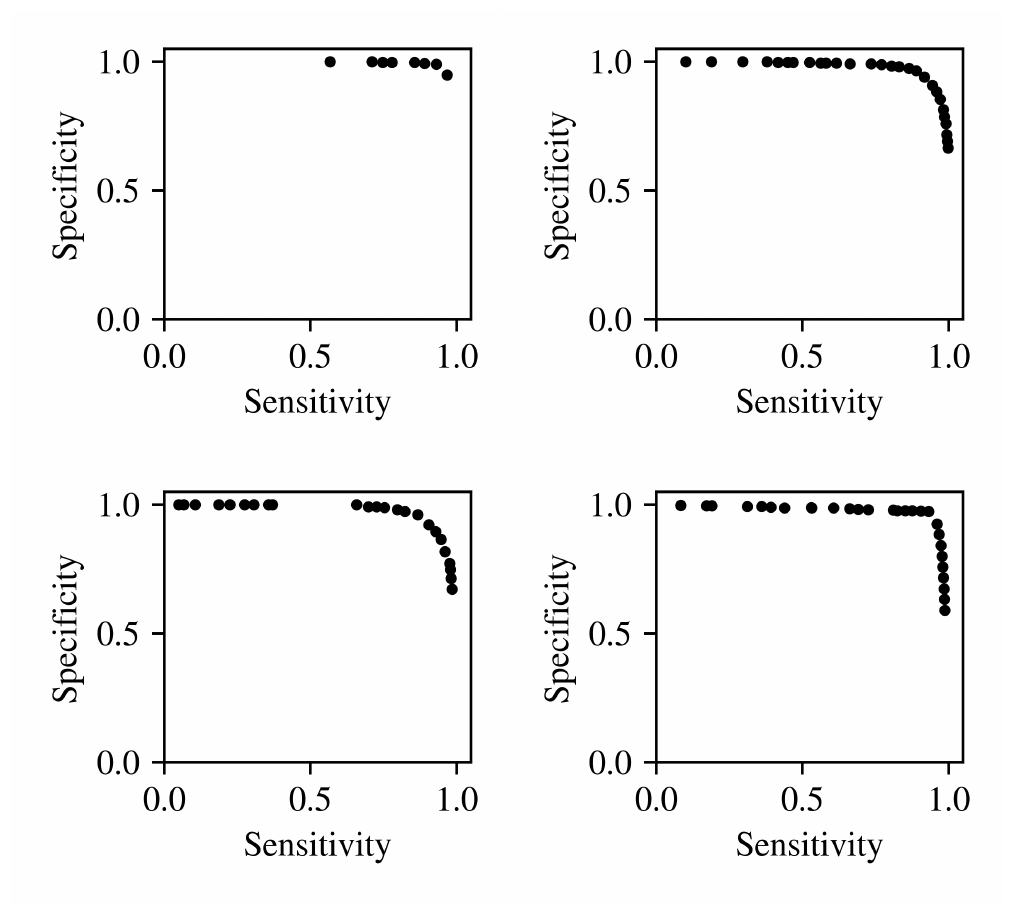}}
    \quad
    \subfloat[]{\includegraphics[width=0.2\textwidth]{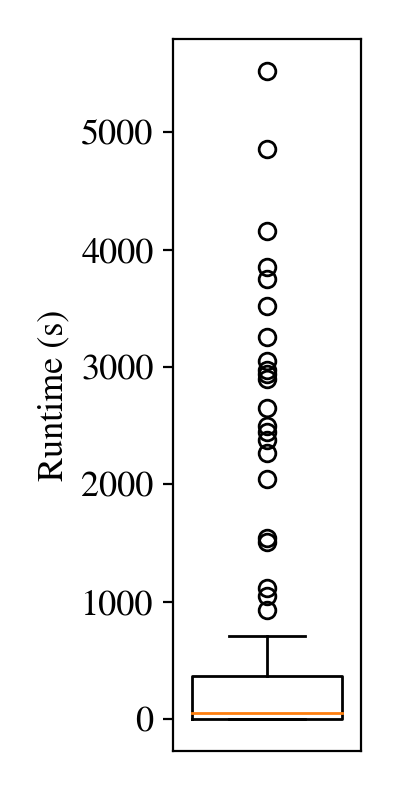}}
    \caption{(a) Examples of sensitivity-specificity trade-off curves for four datasets. (b) Boxplot of the runtimes for generating the sensitivity-specificity trade-off curve.}
    \label{fig:pareto_results}
\end{figure}

As IRELAND generates a pool of AND clauses on the fly, one can readily use these to generate the trade-off curve between sensitivity and specificity. Trade-off curves were generated for the dataset collection with noise. Examples of these trade-off curves are shown in Figure \ref{fig:pareto_results}a, and runtimes are provided in \ref{fig:pareto_results}b.

\section{Discussion}
For large datasets, the primary focus of this work, IRELAND was able to outperform $(BP1)$, the original MILP, both in terms of normalized objective function value and runtime. While $(BP1)$ could not be solved for several instances due to memory issues, IRELAND was able to find a solution for each dataset within 4 hours. For the datasets where $(BP1)$ could not finish within four hours, IRELAND was able to do so and found solutions with an improved normalized objective function value. The column generation approach developed by \cite{dash2018boolean}, called BRCG, is outperformed by IRELAND for datasets without noise. For noisy datasets BRCG largely outperforms IRELAND in terms of runtimes when $J$ is limited. However for large $J$ BRCG often cannot find a solution, while IRELAND is able to do so within four hours.

The number of samples $N$ in a dataset is the best indicator to decide whether to use IRELAND instead of $(BP1)$. For datasets without noise IRELAND always found a solution with an objective function value that was at least as good as the solution found by $(BP1)$, and often much better. Runtimes are improved when the dataset contains more than 1,000 samples for datasets without noise, and for $N\geq 500$ for datasets with noise. When choosing whether to use IRELAND or BRCG the number of features is a good indicator: BRCG gives the best results for datasets up to 1,000 features, while for datasets with more features BRCG often cannot find a solution and IRELAND is the better option. 

IRELAND is similar to a column generation approach in the sense that it consists of a master problem that finds the optimal Boolean phrase in DNF from a given set of AND clauses, and a sub problem that iteratively generates AND clauses that are likely to improve the objective function of the master problem. Directly using column generation has the drawback of a large sub problem when the dataset contains a large number of samples $N$ and features $J$. This was previously observed by \cite{dash2018boolean}, who included a random subset of the features in the sub problem whenever the dataset was large. IRELAND includes the full set of features in the sub problem, but includes only a subset of the samples. This subset of samples is chosen such that all controls are included in order to avoid generating an AND clause that yields a high number of false positives, which would not be used by the master problem. As for the cases, if the random subset of samples would include many cases that were already predicted as cases by the master problem, the newly generated AND clause would have little added value to the set of already existing AND clauses. IRELAND therefore only selects (a subset of) the false negatives, i.e., the cases that were not predicted as cases by the master problem in the previous iteration. This is similar to a column generation approach: one can easily show that in column generation the shadow prices of constraints \eqref{eq:OR1a} are zero when $\hat{y}_n=1$ for $n \in \mathcal{N}_0$.

A binary classification problem is bi-objective by nature: there is a trade-off between the number of true versus the number of false positives. Using the $\varepsilon$-constraint method IRELAND efficiently generates the sensitivity-specificity trade-off curve from the previously generated pool of AND clauses. Using IRELAND it is not necessary to solve multiple large MILPs to generate the trade-off curve. The trade-off between sensitivity and specificity can be very valuable in practical applications, as it allows the user to choose the level of sensitivity or specificity that is most suitable for their application. Note that this trade-off curve is an estimation since (1) IRELAND is a heuristic, (2) the generated trade-off curve depends on the available AND clauses and (3) the upper bounds chosen for solving the sub problems highly impact the granularity of the pool of AND clauses.

IRELAND can handle large datasets of up to 10,000 samples and 10,000 features. Currently dataset sizes are growing rapidly, and in many fields the number of samples and features may grow into the millions. Therefore, though IRELAND is a major step forward, further improvements are needed to keep up with the steady growth of datasets.

When applying classification models it is good practice to split the data into a training, validation and test set. This work however focuses on the development of an algorithm that can efficiently solve the underlying optimization problem during training only, hence no data splitting was used. When applying IRELAND to real-world problems overfitting can be prevented using regularization, for example by tuning the hyperparameters that limit the complexity of the Boolean phrases. Furthermore, the number of true- and false positives resulting from an individual AND clause may be indicative of its potential to overfit: an AND clause that generates only one or a few true positives is less likely to be a true AND clause than one that generates a large number of true positives and a very small number of false positives. As the purpose of this work was to develop a fast optimization approach for the training phase, this is left for future research.

In order to improve generalization one is often interested in Boolean phrases that are as simple as possible, i.e. with a minimum number of AND clauses and a minimum number of features per clause. Currently IRELAND uses an upper bound the number of clauses and the number of features per clause. It could be interesting for the user to see the trade-off between rule complexity and classification accuracy. IRELAND can be extended to accommodate for this by generating a pool of AND clauses that do not only vary regarding the number of false positives, but also vary in the number of included features. The master problem can then be solved for various bounds on the complexity of the final Boolean phrase, yielding the desired trade-off curve. This extension is left for future work.

\section{Conclusion}
IRELAND is an algorithm that can efficiently generate Boolean phrases in disjunctive normal form from datasets with a large number of samples and features. Making use of parallel computation, it generates a large pool of AND clauses representing various trade-offs between the number of true and false positives. From this pool IRELAND can then efficiently generate the sensitivity-specificity trade-off curve, without the need for solving a large number of computationally heavy mixed integer programs.

\section*{Acknowledgements}{This work was supported by the Netherlands Organization for Scientific
Research (NWO) Veni grant VI.Veni.192.043.%
}

%
%
%


\bibliographystyle{informs2014} 
\bibliography{references} 


\end{document}